\pdfoutput=1

\documentclass[11pt]{article}

\usepackage{times}
\usepackage{latexsym}
\usepackage{graphicx,multirow,array,booktabs}

\usepackage[T1]{fontenc}

\usepackage[utf8]{inputenc}

\usepackage{microtype}

\usepackage{inconsolata}

\usepackage[table,xcdraw,dvipsnames]{xcolor}
\usepackage{graphicx}
\usepackage{url}
\usepackage{amsmath,amssymb}
\usepackage{amsfonts}
\usepackage{mathtools}
\usepackage{caption}
\usepackage{subcaption}
\usepackage{wrapfig}
\usepackage{float}

\usepackage{setspace}

\usepackage{tablefootnote}
\usepackage{booktabs,arydshln}

\usepackage[linesnumbered,ruled]{algorithm2e}
\usepackage{multirow}
\usepackage{multicol}
\usepackage{makecell}
\usepackage{bbm}
\usepackage{pifont}

\usepackage[normalem]{ulem}

\usepackage{tocloft}
\usepackage{color,soul}
\usepackage{hyperref}
\usepackage{url}
\hypersetup{
    colorlinks=true,
    linkcolor=blue,
    citecolor=blue
}

\usepackage[shortlabels]{enumitem}
\setlist[itemize]{leftmargin=*}
\setlist[enumerate]{leftmargin=*}
\newcommand{\simulator}{\texttt{DOROTHIE}}
\newcommand{\dataset}{\texttt{SDN}}

\makeatletter
\def\adl@drawiv#1#2#3{%
        \hskip.5\tabcolsep
        \xleaders#3{#2.5\@tempdimb #1{1}#2.5\@tempdimb}%
                #2\z@ plus1fil minus1fil\relax
        \hskip.5\tabcolsep}
\newcommand{\cdashlinelr}[1]{%
  \noalign{\vskip\aboverulesep
           \global\let\@dashdrawstore\adl@draw
           \global\let\adl@draw\adl@drawiv}
  \cdashline{#1}
  \noalign{\global\let\adl@draw\@dashdrawstore
           \vskip\belowrulesep}}
\makeatother

\usepackage{EMNLP2022}

\title{\simulator: Spoken Dialogue for Handling Unexpected Situations \\ in Interactive Autonomous Driving Agents}

\author{
  Ziqiao Ma\textsuperscript{\rm 1}, Ben VanDerPloeg\thanks{\enspace Equal contribution.}\;\textsuperscript{\rm 1}, Cristian-Paul Bara\footnotemark[1]\;\thanks{\enspace Work done prior to joining Amazon Alexa AI}\;\textsuperscript{\rm 1}, Yidong Huang\footnotemark[1]\;\textsuperscript{\rm 1}, \\\textbf{Eui-In Kim\textsuperscript{\rm 1}, Felix Gervits\textsuperscript{\rm 2}, Matthew Marge\textsuperscript{\rm 2}, Joyce Chai\textsuperscript{\rm 1}} \\
  \textsuperscript{\rm 1}University of Michigan \quad \textsuperscript{\rm 2}U.S. Army Research Laboratory \\
  \texttt{\{marstin,bensvdp,cpbara,owenhji,euiink,chaijy\}@umich.edu} \\ \texttt{\{felix.gervits,matthew.r.marge\}.civ@army.mil} \\
}

\date{}

\begin{document}
\maketitle


\begin{abstract}
    In the real world, autonomous driving agents navigate in highly dynamic environments full of unexpected situations where pre-trained models are unreliable.  
    In these situations, what is immediately available to vehicles is often only human operators.
    Empowering autonomous driving agents with the ability to navigate in a \textit{continuous} and \textit{dynamic} environment and to communicate with humans through \textit{sensorimotor-grounded dialogue} becomes critical.  
    To this end, we introduce Dialogue On the ROad To Handle Irregular Events (\simulator), a novel interactive simulation platform that enables the creation of unexpected situations on the fly to support empirical studies on situated communication with autonomous driving agents. 
    Based on this platform, we created the Situated Dialogue Navigation (\dataset), a navigation benchmark of 183 trials with a total of 8415 utterances, around 18.7 hours of control streams and 2.9 hours of trimmed audio.
    \dataset~is developed to evaluate the agent's ability to predict dialogue moves from humans as well as generate its own dialogue moves and physical navigation actions. 
    We further developed a transformer-based baseline model for these \dataset~tasks. 
    Our empirical results indicate that language guided-navigation in a highly dynamic environment is an extremely difficult task for end-to-end models. 
    These results will provide insight towards future work on robust autonomous driving agents\footnote{ The \simulator~platform, \dataset~benchmark, and code for the baseline model are available at \url{https://github.com/sled-group/DOROTHIE}}.
\end{abstract}


\section{Introduction}
\label{sec:intro}

\begin{table*}[!htp]
\hspace*{-0.1cm}
\scalebox{0.51}{
    \begin{tabular}{@{}lccccccccccccccc@{}}
    \toprule
                                &                                   & \multicolumn{2}{c}{\textbf{Environment}}      & \multicolumn{2}{c}{\textbf{Communication}}                         & \multicolumn{2}{c}{\cellcolor[HTML]{FFFFFF}\textbf{Granularity}}                     & \multicolumn{3}{c}{\textbf{Data Collection}}                                                                        & \multicolumn{4}{c}{\textbf{Instruction Type}}    & \textbf{Action Space} \\
\multirow{-2}{*}{\textbf{Name}} & \multirow{-2}{*}{\textbf{Domain}} & \textbf{Fidelity} & \textbf{Continuity}       & \textbf{Turn}             & \textbf{Form}                     & \cellcolor[HTML]{FFFFFF}\textbf{Language} & \cellcolor[HTML]{FFFFFF}\textbf{Control} & \textbf{Lang.}           & \cellcolor[HTML]{FFFFFF}\textbf{Demo.} & \textbf{Modal.}          & \textbf{Replan.}          & \textbf{Adp.} & \textbf{Nav.} & \textbf{Man.}  & \textbf{Continuity}   \\
\cmidrule(lr){1-1} \cmidrule(lr){2-2} \cmidrule(lr){3-4} \cmidrule(lr){5-6} \cmidrule(lr){7-8} \cmidrule(lr){9-11} \cmidrule(lr){12-15} \cmidrule(lr){16-16}
\textbf{\dataset~(Ours)}                   &                                   & Sim               & \cellcolor[HTML]{FFF2CC}C & \cellcolor[HTML]{FFF2CC}M & \cellcolor[HTML]{FFF2CC}Freeform & \cellcolor[HTML]{FFF2CC}H \& L       & \cellcolor[HTML]{FFF2CC}H \& L      & \cellcolor[HTML]{FFF2CC}H   & \cellcolor[HTML]{FFF2CC}H                      & \cellcolor[HTML]{FFF2CC}LVM\textbf{S} & \ding{52} & \ding{52} & \ding{52}    & -         & D \& C               \\
\cmidrule(lr){1-1} \cmidrule(lr){3-4} \cmidrule(lr){5-6} \cmidrule(lr){7-8} \cmidrule(lr){9-11} \cmidrule(lr){12-15} \cmidrule(lr){16-16}
CDNLI~\citep{roh2020conditional}                           &                                   & Sim               & \cellcolor[HTML]{FFF2CC}C & \cellcolor[HTML]{FFF2CC}M & Multi Inst                        &  L       & \cellcolor[HTML]{FFF2CC}H \& L      & \cellcolor[HTML]{FFF2CC}H+T & P                      & LVM                           & - & \ding{52} & \ding{52}    & -    & D \& C                     \\
LCSD~\citep{sriram2019talk}                            &                                   & Sim               & \cellcolor[HTML]{FFF2CC}C & S                         & Multi Inst                        & L                                       & H                                     & \cellcolor[HTML]{FFF2CC}H   & P                                              & LVM                          & - & -     & \ding{52}    & -    & D                     \\
TtW~\citep{de2018talk}                             &                                   & Pano              & D                         & \cellcolor[HTML]{FFF2CC}M & \cellcolor[HTML]{FFF2CC}Freeform & \cellcolor[HTML]{FFF2CC}H \& L       & H                                     & \cellcolor[HTML]{FFF2CC}H   & \cellcolor[HTML]{FFF2CC}H                      & LVM                          & - & -     & \ding{52}    & -    & D                     \\
Talk2Nav~\citep{vasudevan2021talk2nav}                        &                                   & Pano              & D                         & S                         & Multi Inst                        &  L       & H                                     & \cellcolor[HTML]{FFF2CC}H   & P                                              & LVM                           & - & -     & \ding{52}    & -     & D                     \\
TouchDown~\citep{chen2019touchdown}                       &                                   & Pano              & D                         & S                         & Multi Inst                        & L                                       & H                                     & \cellcolor[HTML]{FFF2CC}H   & P                                              & LVM                           & - & -     & \ding{52}    & -      & D                     \\
Street Nav~\citep{hermann2020learning}                      &                                   & Pano              & D                         & \cellcolor[HTML]{FFF2CC}M & Multi Inst                        & L                                       & H                                     & T                           & P                                              & LVM                           & - & -     & \ding{52}    & -    & D                     \\
Map2Seq~\citep{schumann2021generating}                         &                                   & Pano              & D                         & S                         & Multi Inst                        & L                                       & H                                     & \cellcolor[HTML]{FFF2CC}H   & P                                              & LM                           & - & -     & \ding{52}    & -    & D                     \\
RUN~\citep{paz2019run}                             & \multirow{-9}{*}{Outdoors}        & Pano              & D                         & S                         & Multi Inst                        & L                                       & H                                     & \cellcolor[HTML]{FFF2CC}H   & \cellcolor[HTML]{FFF2CC}H                      & LM                           & - & -     & \ding{52}    & -   & D                     \\
    \cmidrule(lr){1-1} \cmidrule(lr){2-2} \cmidrule(lr){3-4} \cmidrule(lr){5-6} \cmidrule(lr){7-8} \cmidrule(lr){9-11} \cmidrule(lr){12-15} \cmidrule(lr){16-16}
TEACh~\citep{padmakumar2022teach}                           &                                   & Sim               & \cellcolor[HTML]{FFF2CC}C & \cellcolor[HTML]{FFF2CC}M & \cellcolor[HTML]{FFF2CC}Freeform & \cellcolor[HTML]{FFF2CC}H \& L       & H                                     & \cellcolor[HTML]{FFF2CC}H   & \cellcolor[HTML]{FFF2CC}H                      & LV                           & -  & \ding{52}  &  \ding{52}    & \ding{52} & D                     \\
DialFRED~\citep{gao2022dialfred}                        &                                   & Sim               & \cellcolor[HTML]{FFF2CC}C & \cellcolor[HTML]{FFF2CC}M & Restricted                         & \cellcolor[HTML]{FFF2CC}H \& L       & H                                     & \cellcolor[HTML]{FFF2CC}H+T                           & P                                              & LV                           & - & \ding{52}  & \ding{52}    & \ding{52}    & D                     \\
ALFRED~\citep{shridhar2020alfred}                          &                                   & Sim               & \cellcolor[HTML]{FFF2CC}C & S                         & Multi Inst                        & \cellcolor[HTML]{FFF2CC}H \& L       & H                                     & \cellcolor[HTML]{FFF2CC}H   & P                                              & LV                           & - & \ding{52}  & \ding{52}    & \ding{52}      & D                     \\
HANNA~\citep{nguyen2019help}                           &                                   & Pano              & D                         & \cellcolor[HTML]{FFF2CC}M & Multi Inst                        & \cellcolor[HTML]{FFF2CC}H \& L       & H                                     & \cellcolor[HTML]{FFF2CC}H   & P                                              & LV                           & - & \ding{52} & \ding{52}    & -    & D                     \\
RobotSlang~\citep{Banerjee2020TheRB}                      &                                   & Phy               & \cellcolor[HTML]{FFF2CC}C & \cellcolor[HTML]{FFF2CC}M & \cellcolor[HTML]{FFF2CC}Freeform & \cellcolor[HTML]{FFF2CC}H \& L                                       & H                                     & \cellcolor[HTML]{FFF2CC}H   & P                                              & LV                           & - & -     & \ding{52}    & -    & D                     \\
TtT and WtW~\citep{ilyevsky2021talk}                       &                                   & Phy               & \cellcolor[HTML]{FFF2CC}C & S                         & Restricted                         & \cellcolor[HTML]{FFF2CC}H \& L       & H                                     & \cellcolor[HTML]{FFF2CC}H   & P                                              & LM                           & - & -     & \ding{52}    & -       & D                     \\
Robo-VLN~\citep{irshad2021hierarchical}                        &                                   & Pano              & \cellcolor[HTML]{FFF2CC}C & S                         & Multi Inst                        & L                                       & \cellcolor[HTML]{FFF2CC}H \& L      & \cellcolor[HTML]{FFF2CC}H   & P                                              & LV                           & - & -     & \ding{52}    & -     & C                     \\
VLN-CE~\citep{krantz2020beyond}                          &                                   & Pano              & \cellcolor[HTML]{FFF2CC}C & S                         & Multi Inst                        & L                                       & H                                     & \cellcolor[HTML]{FFF2CC}H   & P                                              & LV                           & - & -     & \ding{52}    & -     & D                     \\
CVDN~\citep{thomason2020vision}                            &                                   & Pano              & D                         & \cellcolor[HTML]{FFF2CC}M & Restricted                         & L                                       & H                                     & \cellcolor[HTML]{FFF2CC}H   & \cellcolor[HTML]{FFF2CC}H                      & LV                           & - & -     & \ding{52}    & -     & D                     \\
R2R~\citep{anderson2018vision}                             & \multirow{-10}{*}{Indoors}        & Pano              & D                         & S                         & Multi Inst                        & L                                       & H                                     & \cellcolor[HTML]{FFF2CC}H   & P                                              & LV                           & - & -     & \ding{52}    & -    & D     \\
    \bottomrule
\end{tabular}}
\vspace*{0.1cm}
\caption{\small{Comparison of language-conditioned task completion settings in terms of \textbf{Environment Fidelity} (\uline{Sim}ulated, \uline{Pano}ramic, \uline{Phy}sical), \textbf{Environment Continuity} (\uline{D}iscrete, \uline{C}ontinuous), \textbf{Turns of Communication} (\uline{S}ingle, \uline{M}ultiple), \textbf{Communication Form} (\uline{Freeform} Dialogue, \uline{Restricted} Dialogue, \uline{Multi}ple \uline{Inst}ructions), \textbf{Language Granularity} (\uline{H}igh: Goal, \uline{L}ow: Step/Movement), \textbf{Control Granularity} (\uline{H}igh: Action, \uline{L}ow: Control), \textbf{\uline{Lang}uage Collection} (\uline{H}uman, \uline{T}emplated), \textbf{\uline{Demo}nstration Collection} (\uline{H}uman, \uline{P}lanner), \textbf{\uline{Modal}ities} (\uline{L}anguage, \uline{V}ision, \uline{M}ap, \uline{S}peech), \textbf{Instruction Type} (\uline{Replan}ning, \uline{Ad}a\uline{p}tation,  \uline{Nav}igation, \uline{Mani}pulation), \textbf{Action Space} (\uline{D}iscrete, \uline{C}ontinuous)}.}
\label{tab:benchmarks}
\vspace*{-0.5cm}
\end{table*}

\vspace*{-0.25cm}

In embodied agents such as autonomous vehicles (AVs), highly dynamic environments often lead to unexpected situations, such as challenging environment conditions (\textit{e.g.,} caused by weather, light, obstacles, etc.), influence of other agents, and change of the original goals.  
In these situations, the agent's pre-trained models or existing knowledge may not be adequate or reliable to make a corresponding decision. 
What is immediately available to help the agent is often only human partners~\citep{ramachandran2013driver}.  
As they are not programmers who can readily change the code in the field, approaches that enable natural communication and collaboration between humans and autonomy become critical~\citep{spiliotopoulos2001human,weng2016conversational}. 
Although recent years have seen an increasing amount of work in natural language communication with robots, and especially the many benchmarks that have been developed for navigation by instruction following~\citep{roh2020conditional,vasudevan2021talk2nav,shridhar2020alfred,padmakumar2022teach}, little work has been done to study language communication under unexpected situations, particularly in the context of AVs.

To address this limitation, we have developed Dialogue On the ROad To Handle Irregular Events (\simulator), an interactive simulation platform built upon the \texttt{CARLA} simulator~\citep{Dosovitskiy17} to specifically target unexpected situations.  
The \simulator~simulator supports Wizard-of-Oz (WoZ) studies through {\bf a novel duo-wizard setup}: a collaborative wizard (Co-Wizard) that collaborates with the human to accomplish the tasks, and an adversarial wizard (Ad-Wizard) that generates unexpected situations (\textit{e.g.,} creating road obstacles, changing weather conditions, adding/changing goals, etc.) on the fly.  
Using \simulator, we collected the Situated Dialogue Navigation (\dataset) dataset of 183 trials between a Co-Wizard and human subjects to collaboratively resolve unexpected situations and complete navigation tasks through spoken dialogue.

The \dataset~dataset contains multi-faceted and time-synchronized information (\textit{e.g.,} first-person view of the environment, speech input from the human, discrete actions, continuous trajectory and control signals) as well as fine-grained annotation of dialogue phenomena at multiple levels.
\dataset~challenges autonomous driving agents to navigate in continuous and dynamic environments, engage in situated communication with humans, and handle unexpected events on the fly. 
As an initial step, we developed the Temporally-Ordered Task Oriented Transformer (\texttt{TOTO}), a transformer-based baseline model for three tasks: 
(1) predicting dialogue moves from human utterances; 
(2) generating dialogue moves in response to humans; and (3) generating navigation actions towards the goal. 
We present our empirical results and discuss key challenges and opportunities.

To the best of our knowledge, this is the first effort on language communication under unexpected situations in autonomous vehicles. 
Our contributions are the following: 
(1) a novel, high-fidelity simulation platform, \simulator,~that can be used to create unexpected situations on the fly during human-agent communication, 
(2) a fine-grained benchmark,~\dataset,~for continuous, dynamic, interactive navigation with sensorimotor-grounded dialogue, and 
(3) a transformer-based model for action prediction and decision-making which serves as a baseline for future development.  
\section{Related Work}
\label{sec:background}

\begin{figure*}[!htp]
    \centering
    \includegraphics[width=0.9\linewidth]{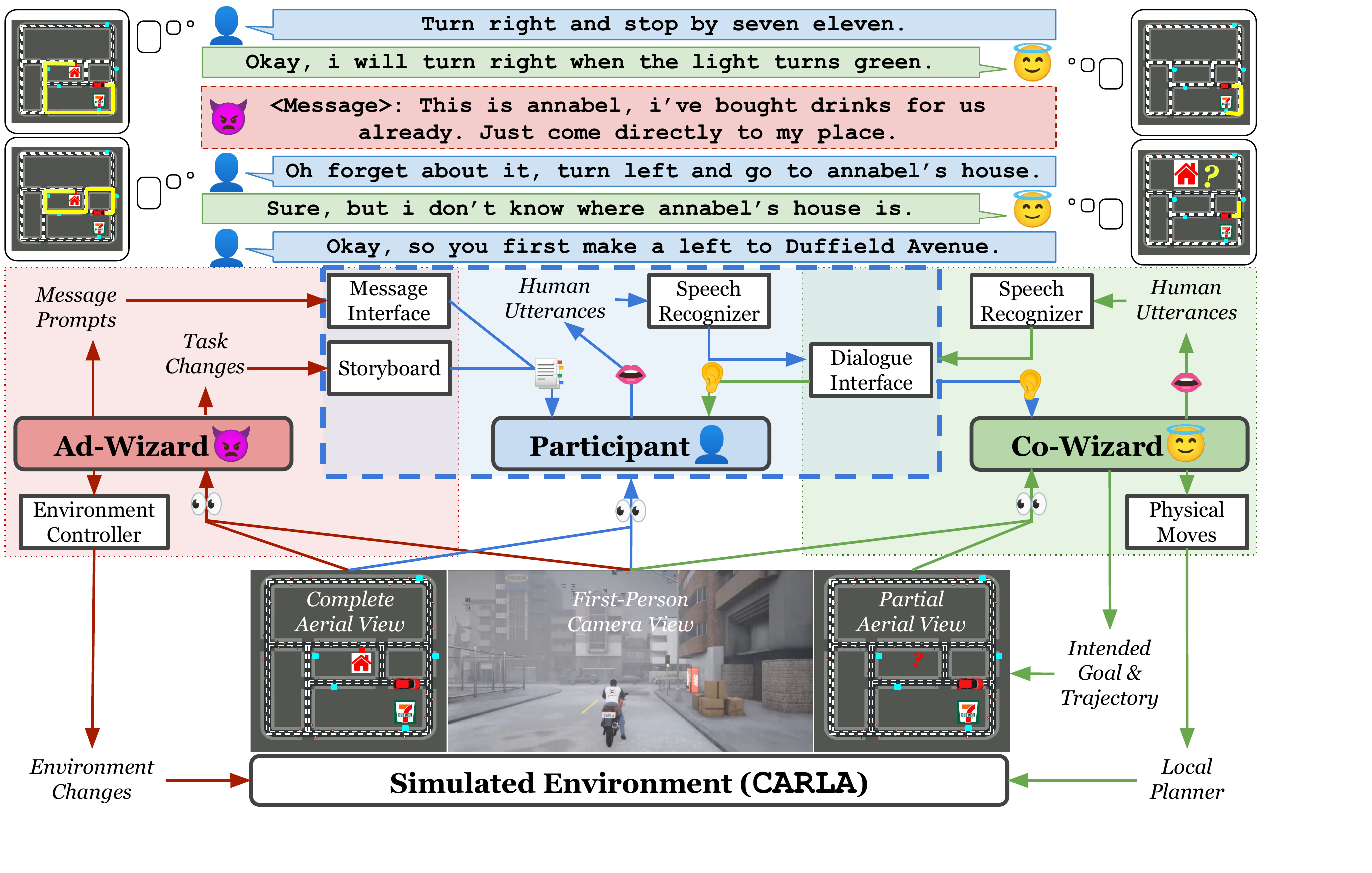}
    \caption{An overview of the \simulator~design. We extend the traditional Wizard-of-Oz framework by introducing a pair of Wizards: \textbf{\textcolor{Green}{Co-Wizard}} and \textbf{\textcolor{Maroon}{Ad-Wizard}}. A Human \textbf{\textcolor{Cyan}{Participant}} is given a storyboard and is instructed to communicate with an autonomous vehicle to complete a set of tasks. The Co-Wizard controls the agent's behaviors and communicates with the human. The Ad-Wizard creates unexpected situations on the fly. The human and the Co-Wizard need to collaborate with each other to resolve these unexpected situations.}
    \label{fig::simulator}
    \vspace*{-0.25cm}
\end{figure*}

Our work is mostly related to \textit{language-conditioned navigation} tasks~\citep{anderson1991hcrc,macmahon2006walk,paz2019run} and particularly recent work on embodied agents that learn to navigate by following language instructions~\cite{gu2022vision}.  
Table~\ref{tab:benchmarks} summarizes the comparison between our work and previous work. 
Below we highlight some key differences.

\paragraph{Replanning in Unexpected Situations.}
Most simulated environments assume that only the tasked agent can change the state of the world through navigation and/or manipulation. 
In outdoor settings, the agent operates in a highly dynamic environment where unexpected changes to the world can often occur due to, \textit{e.g.,} walking pedestrians, moving vehicles, lighting, and weather conditions.  
While previous studies have explored misleading~\citep{roh2020conditional} or perturbed~\citep{lin2021adversarial} instructions, no prior work has looked into how language instructions can help agents adapt in these unexpected situations. 
To our knowledge, \dataset~is the first dataset where language is used to assist agents to replan their goals, paths, and trajectories.

\paragraph{Free-Form Communication.}
Most prior work adopts either simple instruction-following~\citep{chen2019touchdown,shridhar2020alfred,vasudevan2021talk2nav}, or restricted QA dialogue~\citep{chai2018language,thomason2020vision,gao2022dialfred} that only allows the agent to ask for help.
Except for some recent work in human-robot dialogue~\citep{she2017interactive, de2018talk,Banerjee2020TheRB,padmakumar2022teach}, few efforts have supported fully free-form communication where agents can ask, propose, explain, and negotiate under ambiguity or confusion.
To the best of our knowledge, \dataset~is the first benchmark to enable navigation in autonomous driving agents conditioned on free-form spoken dialogue.

\paragraph{Continuous Navigation.}
In discrete navigation, agents take discrete actions, \textit{e.g.,} tele-transport in a pre-defined grid world~\citep{de2018talk} or a navigation graph with sparsely sampled panoramas at each node~\citep{chen2019touchdown,vasudevan2021talk2nav}.
More recently, researchers proposed a continuous navigation setting~\citep{krantz2020beyond,hong2022bridging} by converting discrete paths on navigation graphs into trajectories. 
Unfortunately, these agents are still limited with a discrete action space such as \texttt{forward 0.25m}.
This becomes unnatural in outdoor settings because the default behaviour of outdoor driving agents (\textit{e.g.,} autonomous vehicles) is lane-following instead of staying still.
We instead follow the settings of mobile robot navigation~\citep{roh2020conditional,irshad2021hierarchical}, where the agents are controlled by a continuous action space with physics like throttle and steering, leading to continuous control signals with long-range trajectories.

\section{Dialogue On the ROad To Handle Irregular Events (\simulator) Simulator}
\label{sec:simulator}

Motivated by the wide availability of software simulations for autonomous vehicles~\citep{rosique2019systematic}, we set up our experiment in \texttt{CARLA}~\citep{Dosovitskiy17}, a driving simulator for autonomous vehicles. 
We developed a novel framework, Dialogue On the ROad To Handle Irregular Events (\simulator) (as shown in Figure~\ref{fig::simulator}), to study situated communication under unexpected situations based on the {\em Wizard-of-Oz} (WoZ) paradigm~\citep{riek2012wizard,kawaguchi2004ciair,hansen2005cu,eric2017key}.  
In WoZ, a human participant is typically instructed to interact with an autonomous agent to complete a set of tasks. The agent's behaviors, however, are controlled by a human ``wizard'' (\textit{i.e.,} a researcher).

One important novelty of our framework is that it extends the traditional WoZ approach by introducing a pair of wizards.
In our {\bf duo-wizard} setup, a {\bf Co-Wizard} controls the agent's behaviors and carries out language communication with the human participant to jointly achieve a goal, and an {\bf Ad-Wizard} creates unexpected situations on the fly. 
The Co-Wizard and the participant need to resolve these unexpected situations as they arise.

\subsection{Interface for Co-Wizard Activities}
\label{sec:cowiz}

We found in pilot studies that a low-level, free-form controller is not desirable due to the poor quality of demonstrated trajectories and high cognitive load on the Co-Wizard.
In line with prior work~\citep{roh2020conditional,codevilla2018end,muller2018driving}, we developed a set of high-level {\em physical actions} from pilot studies for the Co-Wizard to control the vehicle.
Each action is mapped to a rule-based local trajectory planner to generate a list of waypoints that the vehicle will drive through.
The continuous control (steering, throttle, brake) of the vehicle is performed by a PID controller.

In a complex navigation task with multiple subgoals, {\em belief tracking} over plans, goals, task status, and knowledge becomes crucial~\citep{ma2012landmark,misu2014situated}.
Besides controlling the vehicle and communicating with the participant, the Co-Wizard also annotates the intended actions (referred to as {\em mental actions}) during and after the interaction, \textit{e.g.}, by noting down the navigation plan by clicking junctions on the intended trajectory from current position to the destination.
The set of the physical and mental actions is described in Figure~\ref{tab:action} and more implementation details are available in Appendix~\ref{app:cowiz-interface}.

\begin{table}[H]
    \centering
    \scalebox{0.5}{
    \begin{tabular}{lll}
            \toprule
            \textbf{Physical Actions} & \textbf{Args}      & \textbf{Descriptions}                 \\
            \midrule
            \texttt{LaneFollow}                & -                      & Default behaviour, follow the current lane. \\
            \texttt{LaneSwitch}                & Angle (\texttt{Rotation}) & Switch to a neighboring lane.               \\
            \texttt{JTurn}                     & Angle (\texttt{Rotation})                & Turn to a connecting road at a junction.    \\
            \texttt{UTurn}                     & -                      & Make a U-turn to the opposite direction.    \\
            \texttt{Stop}                      & -                      & Brake the vehicle manually.                \\
            \texttt{Start}                     & -                      & Start the vehicle manually.                \\
            \cdashlinelr{1-3}
            \texttt{SpeedChange}               & Speed ($\pm$5)        & Change the desired cruise speed by 5 km/h.  \\
            \texttt{LightChange}               & Light State (\texttt{On/Off})    & Change the front light state.     \\
            \midrule
            \textbf{Mental Actions} & \textbf{Args}       & \textbf{Descriptions}                                         \\
            \midrule
            \texttt{PlanUpdate}              & List[\texttt{Junction ID}]           & Indicate intended trajectory towards a destination.           \\
            \texttt{GoalUpdate}              & List[\texttt{Landmark}]              & Indicate current goal as an intended landmark.                \\
            \texttt{StatusUpdate}            & Tuple[\texttt{Landmark,Status}] & Indicate a change in task status. \\
            \texttt{KnowledgeUpdate}          & \texttt{x,y}                & Guess the location of an unknown landmark.      \\
            \texttt{Other}                   & -                       & Other belief state updates.                                   \\   
            \bottomrule 
    \end{tabular}
    }
    \caption{The space of primitive physical actions and mental actions of the Co-Wizard.}
    \label{tab:action}
\end{table}

\subsection{Interface for Ad-Wizard Activities}
\label{sec:adwiz}

The Ad-Wizard is able to introduce {\em environmental exceptions} and {\em task exceptions}.
\begin{itemize}
    \setlength\itemsep{-0.25em}
    \item \textbf{Environmental Exceptions}: Triggered by changes to the environment. These include direct environmental changes, which challenge the vehicle's perceptual processing and motivate participants to request for adaptations without changing the plan or goal (\textit{e.g.}, drive slowly in foggy weather and turn the headlights on at night). Environmental exceptions can also be introduced by creating roadblocks, which motivate new plans by blocking the original ones. 
    \item \textbf{Task Exceptions}: Brought by changing the tasks specified in the storyboard by deleting, adding, or changing a landmark to visit. The Ad-Wizard will send a message to prompt the participant in the message interface with appropriate context, and modify the task interface that specifies the landmarks to visit. Since the Co-Wizard does not have a task interface, the participant needs to communicate with the Co-Wizard in natural language to inform the status of a subgoal, especially when a change of current subgoal is indicated by the Ad-Wizard.
\end{itemize}

The rich dynamics of the environment and tasks in \simulator~create uncertainty and ambiguity, which requires the Co-Wizard to actively initiate conversation with the human partner and find a way to handle these unexpected situations collaboratively.
More illustrated details of the Ad-Wizard interface is available in See Figure~\ref{fig::exception} in Appendix~\ref{app:adwiz-interface}.

\subsection{Data Collection}

Using~\simulator, we recruited 40 na\"ive human subjects as participants for data collection. 
Each subject went through an average of 4.5 sessions. 
In each session, a \textit{storyboard} was given to the subject which required the agent to visit two to six landmarks/destinations.
Each storyboard was generated from four different towns, with all task templates, landmark locations, street names and departure locations randomly shuffled.
While shown the map, the Co-Wizard (an experimenter) did not have access to some of the destinations, \textit{e.g.,} the location of a friend's house or a person to pick up.
Such knowledge disparities motivate rich situated communication and challenge the agent to understand language instructions of different granularity. 
As the Co-Wizard and the human subject communicated with each other to achieve the goal, the Ad-Wizard (another experimenter) was tasked to create different types of unexpected events that were relevant to the current goal. 
The knowledge disparity and unexpected events together drive the communication. 
Details of the task setups are available in Appendix~\ref{app:navigation}.

\section{Situated Dialogue Navigation (\dataset)}
\label{sec:data}

\begin{figure*}[!htp]
    \centering
    \includegraphics[width=1.0\linewidth]{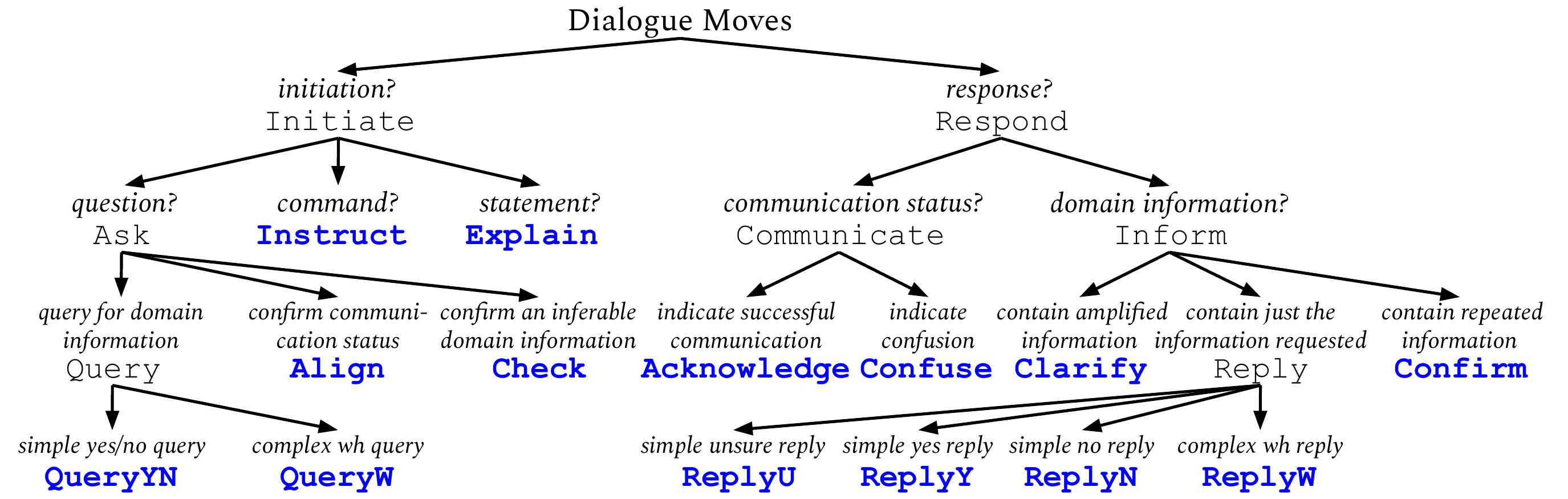}
    \caption{The coding scheme of dialogue moves as a decision tree. The \textcolor{blue}{leaf nodes} of the decision tree specify the set of dialogue moves we used for annotation.}
    \label{fig:dialogue}
    \vspace*{-0.25cm}
\end{figure*}

Our data collection effort has led to the Situated Dialogue Navigation (\dataset), a fine-grained outdoor navigation benchmark.
Each session was replayed at 10 FPS following prior work~\citep{roh2020conditional} to obtain multi-faceted and time-synchronized information, \textit{e.g.,} a first-person view of the environment, speech input from the participant, discrete actions, a continuous trajectory, and control signals.
The benchmark also includes dialogue structure annotation, which we analyzed for dialogue behaviors.

\subsection{Dialogue Structure Annotation}
\label{sec:annotate}

Following prior work in human-robot dialogue~\cite{marge2017exploring,traum2018dialogue,marge2020let} and dialogue discourse processing~\citep{sinclair1975towards,grosz1986attention,clark1996using}, we annotate each dialogue session using four levels of linguistic units:

\begin{itemize}
    \setlength\itemsep{-0.25em}
    \item \textbf{Transaction Units (TUs)}: Sub-dialogues that start when a task is initiated and end when it is completed, interrupted, or abandoned. 
    \item \textbf{Exchange Units (EUs)} Sequences of dialogue moves towards common ground. These start with an initiating utterance that has a purpose (\textit{e.g.,} a question) and end when the expectations are fulfilled or abandoned (\textit{e.g.,} an answer).
    \item \textbf{Dialogue Moves} Sub-categories of dialogue acts that drive conversation and update domain-specific information state within an exchange. 
    \item \textbf{Dialogue Slots} Parameters that further determine the semantics of dialogue moves, including \texttt{Action}, \texttt{Street}, \texttt{Landmark}, \texttt{Status}, \texttt{Object}.
\end{itemize}

We follow the coding scheme of~\citet{carletta1997reliability} to represent dialogue moves as a decision tree, with a slight modification to adjust to our domain, as presented in Figure~\ref{fig:dialogue}. 
The 14 dialogue moves, together with \texttt{Irrelevant}, specify the space of conversational action in the human-vehicle dialogue.
We present an example dialogue with annotations in Figure~\ref{fig::task}, with more samples available in Appendix~\ref{app:sample}.

\subsection{Data Statistics}

\begin{figure}[H]
    \centering
    \vspace*{-0.1cm}
    \begin{subfigure}[t]{.2\textwidth}
        \centering
        \vspace*{0.2cm}
        \scalebox{0.6}{
        \begin{tabular}{cc}
        \toprule
        \textbf{Metric}       & \textbf{Value}       \\
        \midrule
        Control Stream        & 18.7 h        \\
        Trimmed Audio         & 2.9 h          \\
        \# Utterances         & 8415                 \\
        \# Words              & 50398                \\
        Vocabulary            & 1373                 \\
        \# Transactions        & 578                  \\
        \# Exchanges          & 4089                 \\
        \# Dialogue Moves          & 11623                \\
        \# Slot Values        & 8618                 \\
        \# Physical Actions          & 9448                 \\
        \midrule
        \textbf{Fold (Split)} & \textbf{\# Sessions} \\
        \midrule
        Train                 & 123                  \\
        Val (Seen)            & 14                   \\
        Val (Unseen)          & 6                    \\
        Test (Seen)           & 25                   \\
        Test (Unseen)         & 15   \\
        \bottomrule
        \end{tabular}}
    \caption{Dataset Statistics and split information.}
    \label{tab::statistics}
    \end{subfigure}
    ~
    \begin{subfigure}[t]{.26\textwidth}
        \centering
        \vspace*{0.1cm}
        \includegraphics[width=1.0\linewidth]{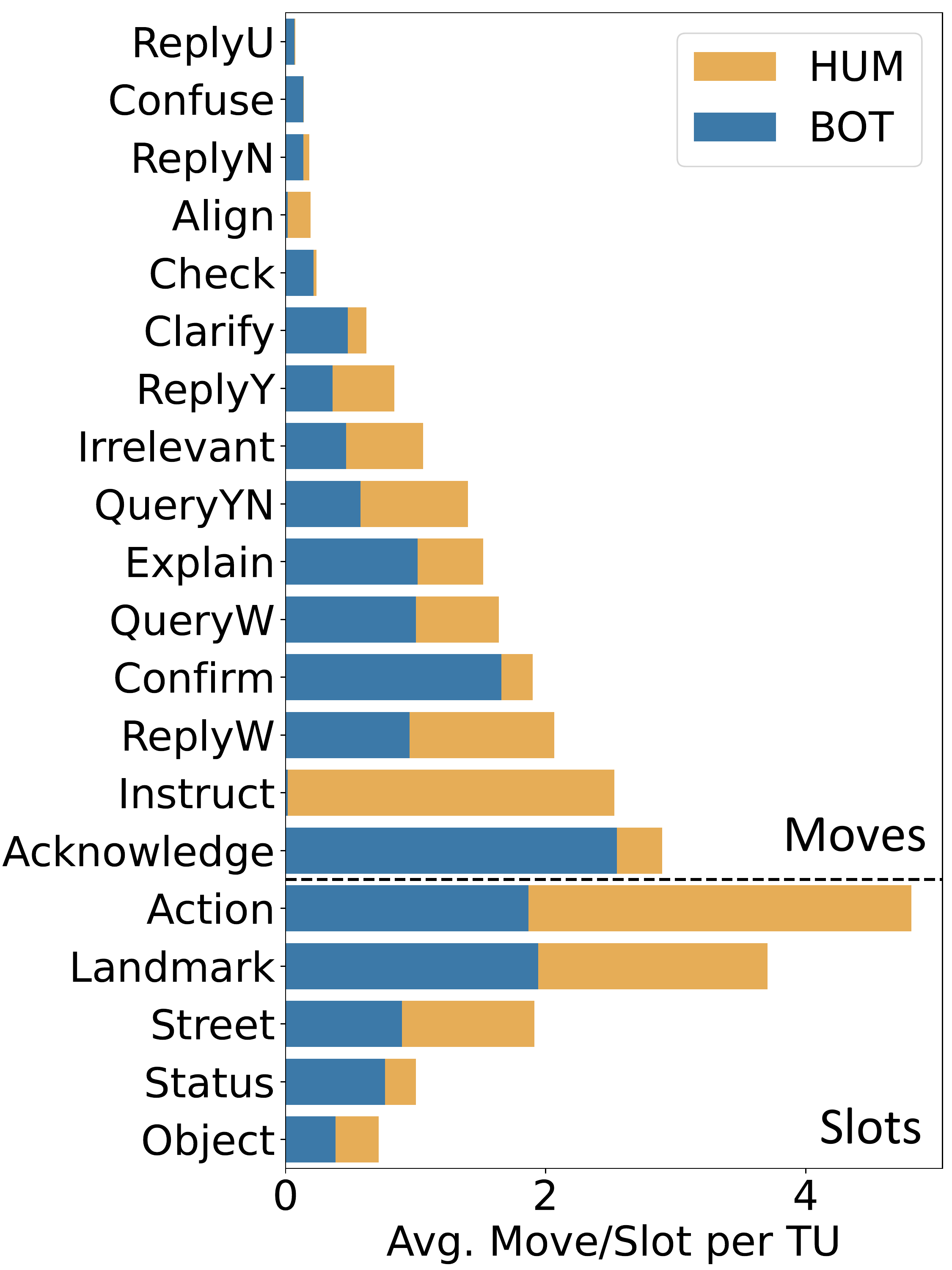}
        \vspace*{-0.57cm}
        \caption{The distribution of dialogue moves and slots per TU.}
        \label{fig::distribution}
    \end{subfigure}
    \vspace*{-0.25cm}
    \caption{Dataset description.}
    \vspace*{-0.1cm}
\end{figure}

The dataset is split into training, validation, and test sets and defines seen (Town 1, 3, 5) and unseen (Town 2) sub-folds for validation and test.
The \dataset~dataset captures rich dialogue behaviors between the human and the agent to collaboratively resolve unexpected situations and achieve joint goals. 
Table~\ref{tab::statistics} shows some basic statistics.


\vspace*{-0.25cm}

Figure~\ref{fig::distribution} shows the frequencies of dialogue moves and slots taken by the human and the agent respectively. 
Not surprisingly, due to the nature of the joint tasks, the human mostly instructs and the agent constantly provides acknowledgement and confirmation. 
Both the human and the agent ask questions and give answers. 
The agent appears to provide more explanation about its own behaviors and decisions.

\begin{figure*}[!htp]
    \centering
      \includegraphics[width=1.0\textwidth]{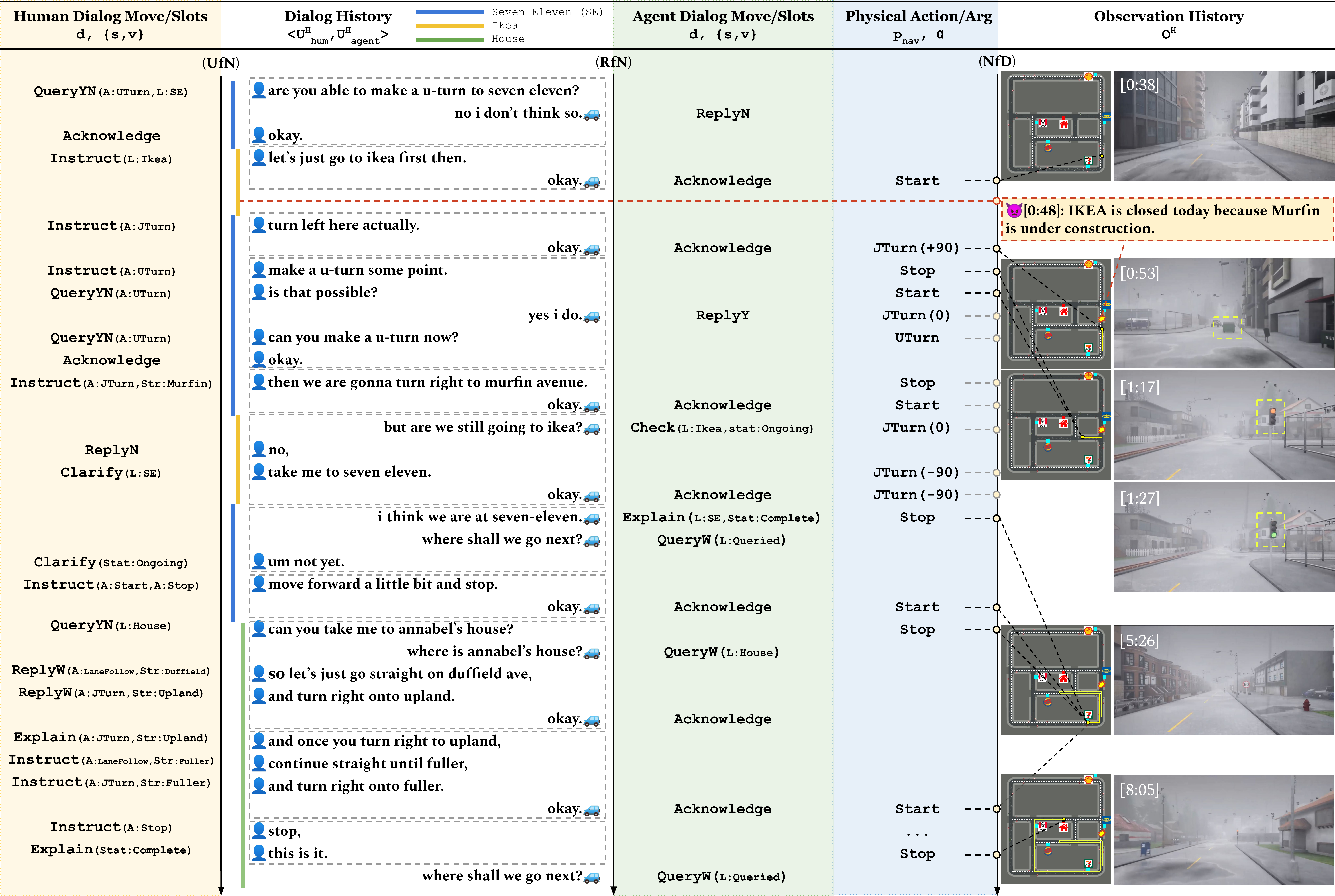}
      \caption{A simple exemplar session in (\dataset) with annotations. Each color bar represents a transaction unit and each box represents an exchange unit. The tasks challenge the agent to understand input dialogue move and imitate Co-Wizard's decision on the next navigation action and dialogue move to take.}
      \label{fig::task}
    \vspace*{-0.2cm}
\end{figure*}

\subsection{Dialogue Behaviors}

\begin{figure*}[!htp]
    \centering
    \begin{subfigure}[t]{.49\textwidth}
        \centering
        \includegraphics[width=1.0\linewidth]{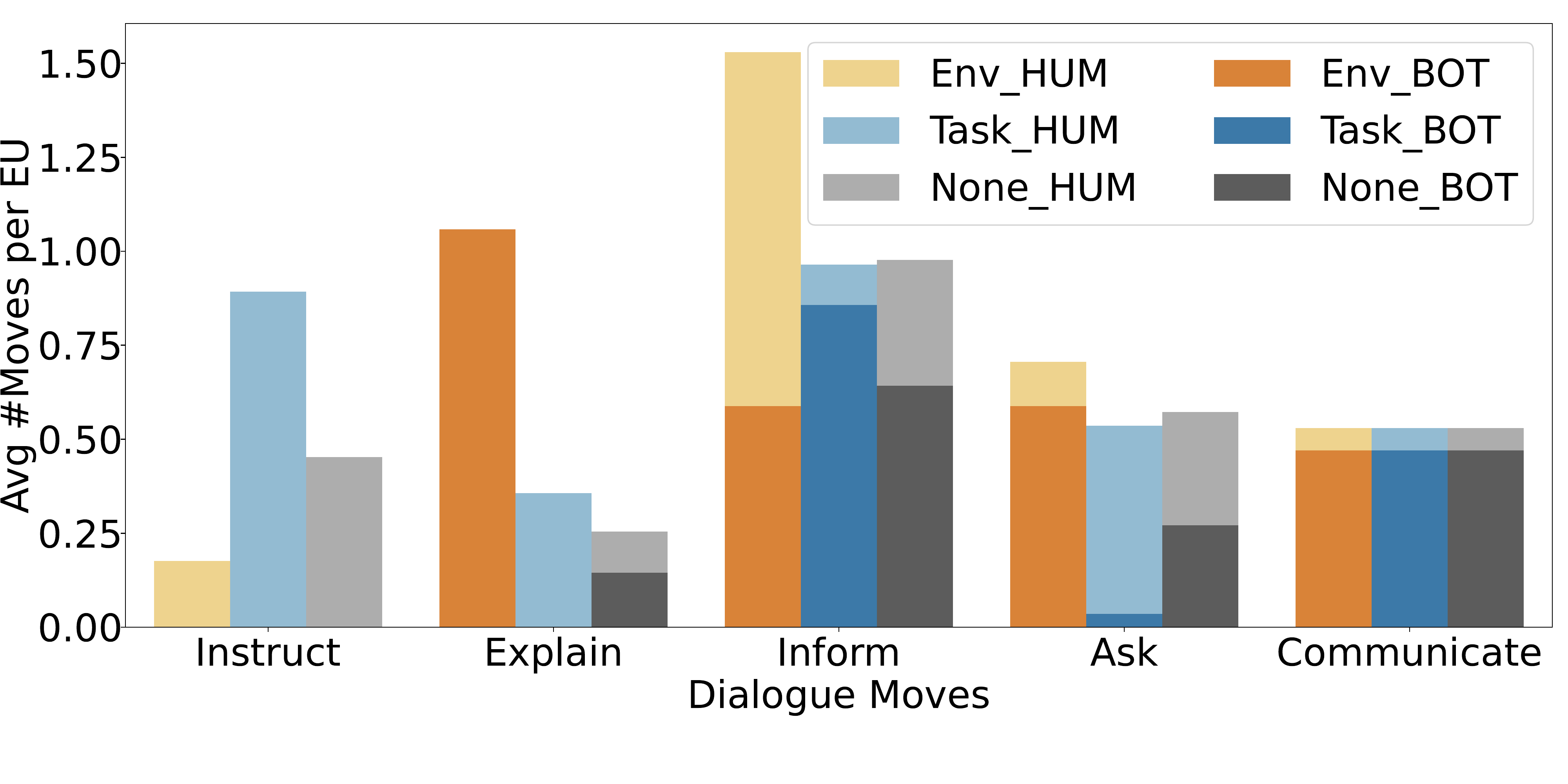}
        \vspace*{-0.75cm}
        \caption{Average \# dialogue moves per EU.}
        \label{fig::distribution-move}
    \end{subfigure}
    ~
    \begin{subfigure}[t]{.49\textwidth}
        \centering
        \includegraphics[width=1.0\linewidth]{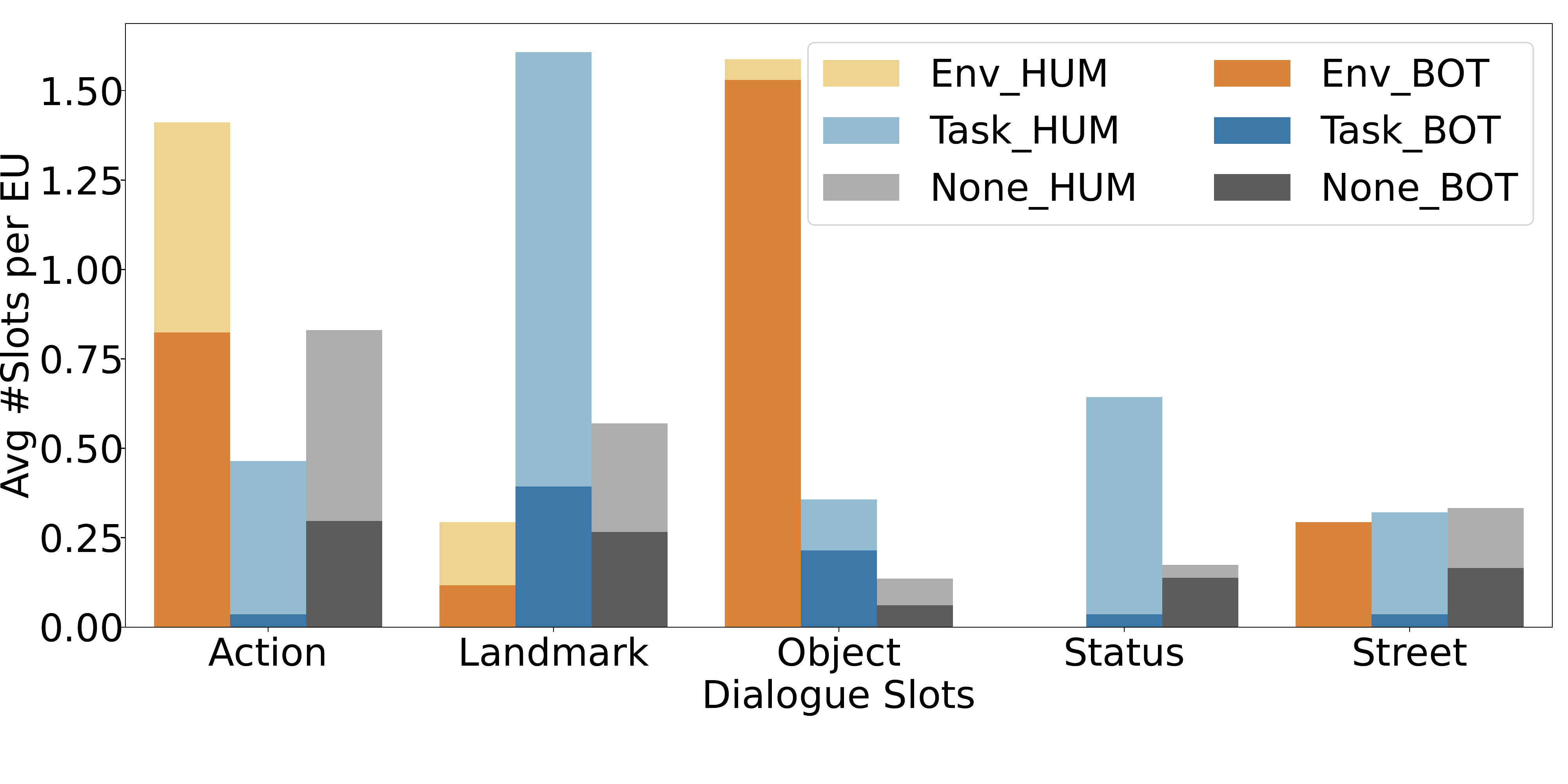}
        \vspace*{-0.75cm}
        \caption{Average \# dialogue slots per EU.}
        \label{fig::distribution-slot}
    \end{subfigure}
    \vspace*{-0.25cm}
    \caption{The average number of dialogue moves and slots per exchange unit (EU). Different colors of the bars categorize the EUs by whether they handle an environmental exception (Env), a task exception (Task), or no exceptions (None). Darker colors indicate the move/slot is produced by the agent (BOT), and lighter colors indicate those produced by human participants (HUM).}
    \label{fig::behaviour}
    \vspace*{-0.5cm}
\end{figure*}

The \dataset~also demonstrates some interesting and unique behaviors between partners to handle unexpected situations. 
In particular, we investigate the distinctive behaviors displayed by the human and the agent to handle different exceptions introduced by the Ad-Wizard.
Figure~\ref{fig::behaviour} shows a comparison of distributions of dialogue moves and slots in EUs.
The EUs are categorized by whether they handle an environmental exception, a task exception, or no exceptions introduced by the Ad-Wizard.
We observe that under environmental exceptions, the agent takes more initiative to describe the situation and ask for help, with frequent use of \texttt{Explain} and \texttt{Ask} moves and use of \texttt{Action} and \texttt{Object} slots. 
In return, the humans initiate less \texttt{Instruct} moves but provide more \texttt{Inform} moves.
Under task exceptions, humans initiate more \texttt{Instruct} moves with frequent use of \texttt{Landmark} and \texttt{Status} slots, in order to describe the change of plan.
The agent makes confirmations with increasing use of \texttt{Inform}.

\section{Task Definition}
\label{sec:task}

While many challenging tasks can be tackled using~\dataset, within the scope of this paper, we formulate three tasks that are critical for enabling situated dialogue for navigation. 
We first introduce some notations then describe our task formulation. 
The agent is provided with domain knowledge $K$, including a list of street names $\{\texttt{str}_i\}$ and (possibly incomplete) landmarks $\{\texttt{lm}_i\}$ on the map topology $M$.
At time $t$, the {\em interaction history}  (possibly empty) is represented as $H_t = \{ O_{t-1}^H, \langle U_{t-1, \texttt{HUM}}^H, U_{t-1, \texttt{BOT}}^H \rangle \}$ which includes visual observations ($O_{t-1}^H$) and dialogue utterances from the human ($U_{t-1, \texttt{HUM}}^H$) and the agent ($U_{t-1, \texttt{BOT}}^H$). The {\em action history}, represented as $A_{t}$, captures the sequence of navigation actions and the dialogue moves from both the human and the agent.  
Given these representations, we define three tasks based on the \dataset~benchmark. 

\paragraph{Dialogue Understanding for Navigation (UfN)}

The UfN task challenges the agent to understand human intention (\textit{i.e.,} dialogue moves) from an incoming utterance. We consider each point in the \dataset~where the human makes an utterance as an {\bf \em inference point} $\tau$. The task is to, at each inference point $\tau$, predict the dialogue move-slots pair $\langle d, \{s,v\} \rangle$ of the incoming utterance $u_\tau$ given the knowledge and history $\{ K, M, H_\tau, A_\tau \}$.

\paragraph{Dialogue Response for Navigation (RfN)}

The RfN task challenges the agent to generate the adequate dialogue move-slot pair to drive communication. We consider each point in~\dataset ~where the Co-Wizard selected a dialogue action and/or navigation action as a {\bf \em decision point}. The task is to, at each decision point $\tau$, generate the dialogue move-slots pair $\langle d, \{s,v\} \rangle$ given the knowledge and history $\{ K, M, H_\tau, A_\tau \}$.

\paragraph{Navigation from Dialogue (NfD)}

The NfD task challenges the agent to follow human instructions from dialogue history.
The task is to, at each decision point $\tau$ for navigation, generate the action-argument pair $\langle p, \alpha \rangle$ for navigation given the knowledge and history $\{ K, M, H_\tau, A_\tau \}$.

\paragraph{Evaluation}

To ensure all unexpected events and future dialogue still make sense, tasks are defined and evaluated in a teacher-forcing manner~\citep{lamb2016professor,anderson2018vision}, \textit{e.g.,} the action history $A_\tau$ presented to the model will always be the ground truth during data collection, instead of those predicted by the model at inference time.
For the UfN and RfN tasks, we report the \textbf{move accuracy} and \textbf{dialogue slot F1-score} of each dialogue move and slot-value pair.
In the NfD task, the argument for navigation actions is a yaw rotation angle $\alpha \in [-180, 180)$.
During evaluation, a prediction that deviates for less than 15 degrees will be considered accurate.
We report the \textbf{action accuracy} with and without argument.

\section{Temporally-Ordered Task-Oriented Transformer (\texttt{TOTO})}

\begin{figure*}[!htp]
    \centering
    \includegraphics[width=1.0\linewidth]{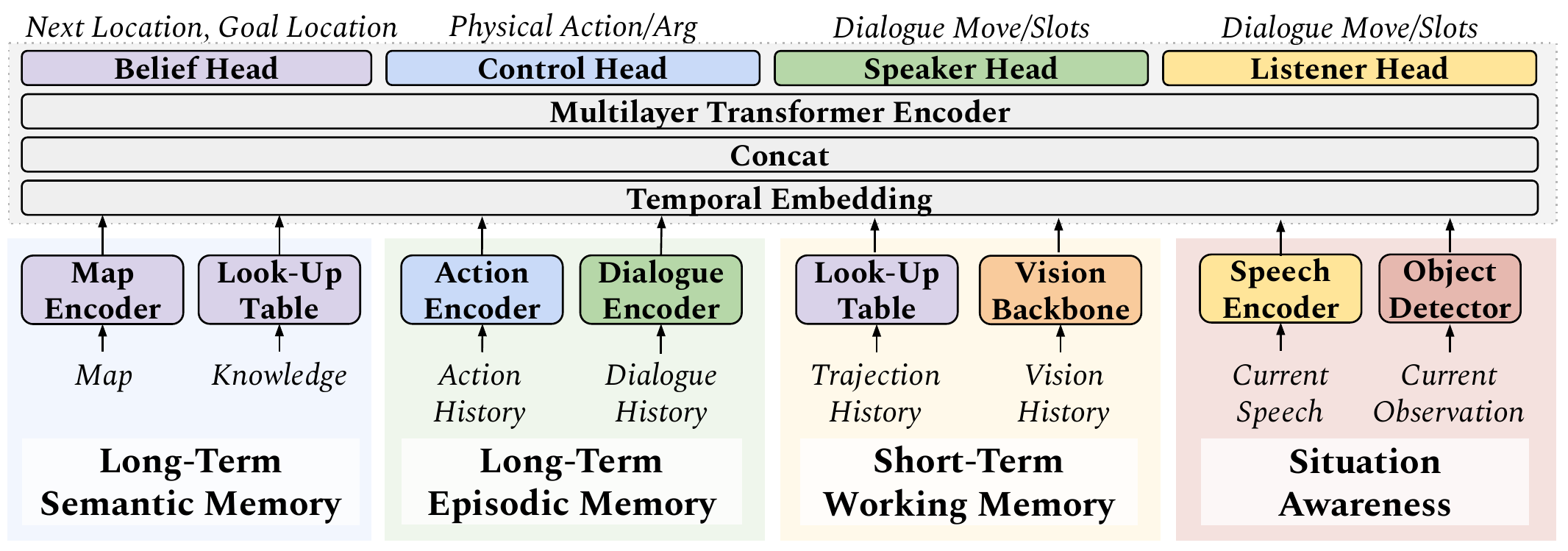}
    \caption{An overview of the architecture of the Temporally-Ordered Task-Oriented (\texttt{TOTO}) Transformer.}
    \label{fig::model}
    \vspace*{-0.5cm}
\end{figure*}

Motivated by recent advances in decision-making transformers~\citep{chen2021decision,pashevich2021episodic,zhang2021hierarchical}, we present Temporally-Ordered Task-Oriented Transformer (\texttt{TOTO}), a Transformer-based baseline.
\texttt{TOTO} is temporally-ordered as it assigns sinusoidal temporal encodings for input history instead of recurrent updates of hidden state, and is task-oriented as a unified architecture for all 3 tasks on the \dataset~benchmark.
The text, speech, and vision inputs are each encoded using frozen pre-trained unimodal models.
After encoding, the temporally encoded input is concatenated and passed through a multi-layer transformer. 
The output embeddings are sent to fully-connected layers to decode task outputs.
The model architecture is illustrated in Figure~\ref{fig::model}, and more details are provided in Appendix~\ref{app:baseline}.

\paragraph{Long-Term Semantic Memory}

For each known landmark, we encode it with a look-up table for its location and a pre-trained BERT model~\citep{devlin2019bert} for its name.
While each street corresponds to a subgraph on the map topology, we first encode the map with a graph attention layer~\citep{velickovic2018graph}, and then concatenate the pooling of each subgraph with the BERT embedding for street names.
Since the knowledge $K$ and map topology $M$ are provided from the beginning of a session, each knowledge and street embedding is assigned with a zero temporal encoding.

\paragraph{Long-Term Episodic Memory}

We encode the complete dialogue history and action history in this module.
The transcribed dialogue history, together with special speaker role tokens, is tokenized and encoded by a pre-trained BERT model.
The full navigation action history, including both navigation actions and dialogue moves with their arguments and slots, is encoded with look-up tables following prior work~\citep{pashevich2021episodic}.
Each utterance and action embedding is assigned with their corresponding temporal encoding.

\paragraph{Short-Term Working Memory}

Given the long-range nature of \dataset, encoding the complete vision and trajectory history is not computationally realistic.
We instead assign a fixed window size of $T = 100$ with a step $\Delta t = 4$ to sample and encode the nearest vision and trajectory history.
We refer to this module as the short-term working memory encoder.
Each location on the trajectory is encoded with the same look-up table in semantic memory, and each image in the visual stream is encoded with a pre-trained ResNet-50~\citep{he2016deep} backbone.

\paragraph{Situation Awareness}

Situational awareness is crucial to handle unexpected events. 
To this end, we attend to the current speech input with a pre-trained HuBERT~\citep{hsu2021hubert} encoder.
In addition, we train a transformer-based object detector from 30k images sampled from seen splits in \texttt{CARLA}.
The model is based on Deformable DETR~\citep{zhu2021deformable} and SegFormer~\citep{xie2021segformer}, pre-trained from the supervision of bounding boxes as well as depth and semantic segmentation obtained from pseudo-sensors.
The speech and object embeddings are assigned with the temporal encoding of current timestamp $t$.

\section{Experiments and Results}
\label{sec:experiment}

\begin{table*}[!htp]
\centering
\scalebox{0.76}{
    \begin{tabular}{lcccccc}
    \toprule
    \multicolumn{1}{c}{\multirow{2}{*}{\textbf{Model}}} & \multicolumn{2}{c}{\textbf{UfN (Seen)}} & \multicolumn{2}{c}{\textbf{RfN (Seen)}} & \multicolumn{2}{c}{\textbf{NfD (Seen)}} \\
    \multicolumn{1}{c}{}                                & \textbf{Move Acc.}    & \textbf{Slot F1}       & \textbf{Move Acc.}    & \textbf{Slot F1}      & \textbf{Action Acc.}       & \textbf{Act-Arg Joint Acc.} \\
   \cmidrule(lr){1-1} \cmidrule(lr){2-3} \cmidrule(lr){4-5} \cmidrule(lr){6-7}
    TOTO                                               
    & $ 40.9 _{(\pm 3.9 )}$ & $ 36.9 _{(\pm  0.0 )}$  & $ 29.2 _{(\pm  0.7 )}$ & $ 55.7 _{(\pm  0.2 )}$ & $ 41.2 _{(\pm  2.5 )}$ & $ 36.0 _{(\pm  3.4 )}$ \\
    \cmidrule(lr){1-1} \cmidrule(lr){2-3} \cmidrule(lr){4-5} \cmidrule(lr){6-7}
    TOTO (+ Belief Tracking)                                   
    & $ 39.5 _{(\pm 2.2 )}$ & $ 37.0 _{(\pm  0.1 )}$  & $ 28.8 _{(\pm  0.9 )}$ & $ 55.7 _{(\pm  0.2 )}$ & $ 40.7 _{(\pm  3.6 )}$ & $ 34.0 _{(\pm  4.7 )}$\\
    TOTO (- Action History)                             
    & \textcolor{red}{$ 30.5 _{(\pm 1.5 )}$} & $ 36.9 _{(\pm  0.0 )}$  & \textcolor{red}{$ 23.5 _{(\pm  1.7 )}$} & $ 55.7 _{(\pm  0.0 )}$ & \textcolor{red}{$ 27.6 _{(\pm  2.8 )}$} & \textcolor{red}{$ 24.6 _{(\pm  4.0 )}$}  \\
    TOTO (- GT Transcript)                             
    & $ 39.8 _{(\pm 1.9 )}$ & $ 36.9 _{(\pm  0.1 )}$  & $ 29.2 _{(\pm  0.8 )}$ & $ 55.6 _{(\pm  0.1 )}$ & $ 40.4 _{(\pm  3.4 )}$ & $ 31.6 _{(\pm  4.3 )}$ \\
    TOTO (- Object Detection)                           
    & $ 42.5 _{(\pm 2.8 )}$ & $ 37.0 _{(\pm  0.2 )}$  & $ 30.4 _{(\pm  0.7 )}$ & $ 55.8 _{(\pm  0.1 )}$ & $ 39.2 _{(\pm  3.5 )}$ & $ 34.4 _{(\pm  5.8 )}$  \\
    TOTO (- Vision History)                             
& $ 41.9 _{(\pm 1.3 )}$ & $ 37.0 _{(\pm  0.2 )}$  & $ 29.1 _{(\pm  0.5 )}$ & $ 55.8 _{(\pm  0.2 )}$ & $ 42.0 _{(\pm  3.1 )}$ & $ 36.1 _{(\pm  4.0 )}$ \\
    TOTO (- Current Speech)                             
& \textcolor{red}{$ 35.1 _{(\pm 2.7 )}$} & $ 36.7 _{(\pm  0.5 )}$  & $ 29.9 _{(\pm  0.9 )}$ & $ 55.9 _{(\pm  0.2 )}$ & $ 39.7 _{(\pm  1.9 )}$ & $ 33.7 _{(\pm  3.0 )}$\\
    TOTO (- Map Knowledge)                             
& $ 42.6 _{(\pm 1.2 )}$ & $ 36.9 _{(\pm  0.0 )}$  & $ 29.3 _{(\pm  0.9 )}$ & $ 55.8 _{(\pm  0.2 )}$ & $ 44.6 _{(\pm  3.3 )}$ & $ 39.1 _{(\pm  3.3 )}$ \\
    \cmidrule(lr){1-1} \cmidrule(lr){2-3} \cmidrule(lr){4-5} \cmidrule(lr){6-7}
    Episodic Transformer
& \textcolor{red}{$ 36.6 _{(\pm 3.6 )}$} & $ 37.0 _{(\pm  0.2 )}$  & $ 29.4 _{(\pm  1.2 )}$ & $ 55.9 _{(\pm  0.2 )}$ & $ 40.0 _{(\pm  2.8 )}$ & $ 32.2 _{(\pm  4.0 )}$\\  
    Fine-tuned BERT 
& $\mathbf{66.8 _{(\pm 2.0 )}}$ & \textcolor{red}{$ 24.9 _{(\pm  5.5 )}$}  & $\mathbf{52.7 _{(\pm  1.0 )}}$ & \textcolor{red}{$ 46.0 _{(\pm  2.5 )}$} & \textcolor{red}{$32.4 _{(\pm  1.2 )}$} & \textcolor{red}{$ 16.2 _{(\pm  2.7 )}$}\\
    \midrule
    \multicolumn{1}{c}{\multirow{2}{*}{\textbf{Model}}} & \multicolumn{2}{c}{\textbf{UfN (Unseen)}} & \multicolumn{2}{c}{\textbf{RfN (Unseen)}} & \multicolumn{2}{c}{\textbf{NfD (Unseen)}} \\
    \multicolumn{1}{c}{}                                & \textbf{Move Acc.}    & \textbf{Slot F1}       & \textbf{Move Acc.}    & \textbf{Slot F1}      & \textbf{Action Acc.}       & \textbf{Act-Arg Joint Acc.} \\
    \cmidrule(lr){1-1} \cmidrule(lr){2-3} \cmidrule(lr){4-5} \cmidrule(lr){6-7}
    TOTO                                            
& $ 49.2 _{(\pm 3.0 )}$ & $ 26.2 _{(\pm  0.0 )}$  & $ 31.0 _{(\pm  1.7 )}$ & $ 54.0 _{(\pm  0.7 )}$ & $ 45.8 _{(\pm  3.8 )}$ & $ 41.1 _{(\pm  2.8 )}$\\
    \cmidrule(lr){1-1} \cmidrule(lr){2-3} \cmidrule(lr){4-5} \cmidrule(lr){6-7}
    TOTO (+ Belief Tracking)                                   
& $ 47.1 _{(\pm 3.5 )}$ & $ 26.2 _{(\pm  0.0 )}$  & $ 29.0 _{(\pm  2.0 )}$ & $ 53.7 _{(\pm  0.7 )}$ & $ 47.6 _{(\pm  4.5 )}$ & $ 38.8 _{(\pm  3.1 )}$\\
    TOTO (- Action History)                             
& \textcolor{red}{$ 35.5 _{(\pm 3.2 )}$} & $ 26.1 _{(\pm  0.1 )}$  & $ 28.2 _{(\pm  3.9 )}$ & $ 54.8 _{(\pm  0.0 )}$ & \textcolor{red}{$ 36.8 _{(\pm  0.8 )}$} & $ 36.0 _{(\pm  1.7 )}$\\
    TOTO (- GT Transcript)                             
& $ 46.7 _{(\pm 2.4 )}$ & $ 26.2 _{(\pm  0.0 )}$  & $ 31.6 _{(\pm  2.6 )}$ & $ 54.2 _{(\pm  0.8 )}$ & $ 46.2 _{(\pm  5.9 )}$ & $ 37.6 _{(\pm  6.9 )}$\\
    TOTO (- Object Detection)                           
& $ 50.0 _{(\pm 1.8 )}$ & $ 26.2 _{(\pm  0.1 )}$  & $ 32.7 _{(\pm  2.2 )}$ & $ 53.8 _{(\pm  1.2 )}$ & $ 45.7 _{(\pm  5.2 )}$ & $ 40.3 _{(\pm  5.4 )}$\\
    TOTO (- Vision History)                             
& $ 48.7 _{(\pm 2.3 )}$ & $ 26.2 _{(\pm  0.1 )}$  & $ 31.5 _{(\pm  2.9 )}$ & $ 54.3 _{(\pm  0.7 )}$ & $ 45.9 _{(\pm  4.2 )}$ & $ 42.3 _{(\pm  3.5 )}$\\
    TOTO (- Current Speech)                             
& \textcolor{red}{$ 42.8 _{(\pm 2.5 )}$} & $ 25.8 _{(\pm  0.3 )}$  & $ 33.8 _{(\pm  1.4 )}$ & $ 55.1 _{(\pm  0.4 )}$ & $ 46.5 _{(\pm  4.9 )}$ & $ 39.4 _{(\pm  5.2 )}$\\
    TOTO (- Map Knowledge)                             
& $ 48.2 _{(\pm 1.0 )}$ & $ 26.2 _{(\pm  0.1 )}$  & $ 31.9 _{(\pm  1.2 )}$ & $ 54.9 _{(\pm  0.8 )}$ & $ 51.7 _{(\pm  3.4 )}$ & $ 46.0 _{(\pm  4.0 )}$\\
    \cmidrule(lr){1-1} \cmidrule(lr){2-3} \cmidrule(lr){4-5} \cmidrule(lr){6-7}
    Episodic Transformer
& $ 45.1 _{(\pm 3.8 )}$ & $ 26.1 _{(\pm  0.1 )}$  & $ 33.4 _{(\pm  2.2 )}$ & $ 54.7 _{(\pm  0.8 )}$ & $ 46.6 _{(\pm  3.3 )}$ & $ 37.0 _{(\pm  5.9 )}$\\
    Fine-tuned BERT 
& $\mathbf{67.2 _{(\pm 1.5 )}}$ & \textcolor{red}{$ 16.2 _{(\pm  3.5 )}$}  & $\mathbf{57.0 _{(\pm  0.9 )}}$ & \textcolor{red}{$ 46.9 _{(\pm  2.2 )}$} & \textcolor{red}{$ 37.1 _{(\pm  1.5 )}$} & \textcolor{red}{$ 19.6 _{(\pm  3.6 )}$} \\
    \bottomrule
    \end{tabular}
}
\vspace*{-0.1cm}
\caption{Experiment results of \texttt{TOTO} and baselines on the three tasks in the \dataset~benchmark. We use accuracy as the primary evaluation metric for navigation actions and dialogue moves, and use F1-score as the primary metric for dialogue slot, both in percentage. Each experiment is repeated with 5 random seeds. In each run, the model is validated on the complete validation fold (including both seen and unseen splits). The model is trained until overfit or reaching the max epochs. The model with lowest loss on the validation fold will be used for inference.}
\label{tab:results}
\vspace*{-0.25cm}
\end{table*}

We summarize the experiment results in Table~\ref{tab:results}.
Our initial end-to-end transformer model is able to handle all tasks uniformly on both the seen and unseen splits of the test set, and outperform the majority of the unimodal baselines (See Table~\ref{tab:results-full} for full results).
In general, the performance is more comparable on inference tasks than decision tasks, \textit{e.g.,} predicting the dialogue moves from human utterances is a more approachable task than predicting the dialogue moves and navigation actions in response.
We also noticed that the results on the unseen splits are uniformly better than the seen set. 
This can partially be explained by the fact that the unseen environment (Town 2) is significantly smaller in size and simpler in map topology.
Overall, our experiment has shown that the tasks in \dataset~are challenging.
Comparatively, the Episodic Transformer (E.T.)~\citep{pashevich2021episodic} baseline particularly underperforms in the inference task, \textit{i.e.}, UfN move prediction.
The fine-tuned language model baseline can handle dialogue move predictions very well, but significantly fails on other tasks.
We further provide a set of ablation studies, and discuss potential reasons why \dataset~is a challenge for end-to-end models.
Additional results are available in Appendix~\ref{app:results}.

\paragraph{Ablation on Input Modalities}
To understand how each input modality contributes to task performance, we conduct ablation studies by removing one of the input modalities.
All experiments on action-level tasks (inferring or predicting dialogue moves or physical actions) are mostly influenced by the action history, which can be explained by the fact that ground truth action history is available in the teacher-forcing setup.
Not surprisingly, the understanding of the incoming utterance is also largely influenced by the input of current speech.
Counter-intuitively, we noticed that removing the map and knowledge encoder does not lead to decreased performance.
This observation suggests that the end-to-end approach may not be reliable in tasks that require reasoning and planning over graphs, especially route planning.
Overall, there is no statistically significant evidence that the full model benefits from perceptual history.
We also notice that the slot-F1 scores stays relatively constant across ablation, indicating that the slot-value prediction remains challenging for \texttt{TOTO}.

\paragraph{Ablation on Belief Tracking} 

Prior work~\citep{ma2019selfmonitoring,zhang2021hierarchical} has indicated the power of task monitoring.
To understand if end-to-end belief tracking would benefit the computational model, we additionally introduce a belief head with auxiliary loss, which is tasked to predict the location of the next timestamp on the trajectory and the location of the goal landmark annotated during data collection.
According to Table~\ref{tab:results}, we observe marginal but no statistically significant improvement.
This observation suggests that for long-range navigation tasks with unexpected goal changes, end-to-end approaches can hardly benefit from end-to-end belief tracking.
Other modeling approaches should be explored to make full use of the rich belief update annotations of \dataset.

\section{Conclusion}
\label{sec:conclusion}

We introduced \simulator, a high-fidelity simulation platform to support WoZ studies for situated communication with autonomous driving agents that can adapt to unexpected events.
We defined and collected \dataset, a fine-grained benchmark for continuous, dynamic, interactive navigation with sensorimotor-grounded dialogue.
Our \simulator~platform, together with our \dataset~benchmark, contribute a valuable resource for several lines of work in Robotics Navigation and Human-Robot Communication. 
We presented Temporally-Order Task Oriented Transformer (\texttt{TOTO}), a fully transformer-based baseline model for the \dataset~task.
Our empirical results have shown that such long-horizon navigation tasks with rich dialogue phenomena and unexpected situations can be very challenging for end-to-end approaches.
This work has shown that language-guided navigation in a highly dynamic environment (\textit{e.g.,} in the context of AVs) is an extremely difficult task. 
Our \simulator~simulation environment, the \dataset~benchmark, and baseline models provide a stepping stone towards future efforts in this challenging space.


\vspace*{-0.1cm}
\section*{Acknowledgements}
This work was supported in part by the Automotive Research Center (ARC) at the University of Michigan, 
NSF IIS-1949634, 
and NSF SES-2128623. 
The authors would like to thank Yichi Zhang and Jiayi Pan for their helpful discussions and the anonymous reviewers for their valuable feedback.

\section*{Limitations}
\label{sec:limitation}

While the \dataset~collects control streams of the vehicle, we scope our preliminary experiment on high-level robotic action prediction due to the complexity of the full problem. 
Specifically, our current task setup only involves the prediction of physical action and dialogue moves, and is based on several assumptions. 
First, there is an oracle that decides when to initiate a physical action or dialogue move. 
Second, we assume a ground truth low-level local planner that maps a navigation action to a sequence of control. 
Third, we bypass the complexity of language generation. 
We also ignore the adaptive and epistemic actions (speed and light changes) involved and focus on navigation. 
These assumptions allow us to focus on the understanding and response of situated dialogue in the navigation task and enable automatic evaluation.

In the future, we will expand the model to fully autonomous navigation settings, with sequential decision-making and hierarchical control policies.
We are also interested in computational approaches for situated dialogue state tracking and management, especially involving in natural language generation and human partner modeling.
With the above milestones accomplished, we will look into Sim2Real transfer and deploy our algorithm in physical autonomous driving agents.

\section*{Ethics Statement}

The institution’s Institutional Review Board (IRB) considered this project exempt from ongoing review.
The \dataset~benchmark contains human generated data (speech and demonstrations).
Due to the Wizard-of-Oz nature of the study, the participants consent preceding the study and are debriefed at the end of the study.
The data collection among research staff and volunteers are in line with standard ethical practice. 
For broader social impact, \simulator~aims at empowering autonomous vehicles with the ability to harness human knowledge and expertise through dialogue, and enabling natural language communication and collaboration in tackling unexpected situations.
Since the dataset was developed from the simulator, the safety concerns are minimal. 
A complete ethical statement is available in Appendix~\ref{app:ethics}.


\bibliography{reference}
\bibliographystyle{acl_natbib}


\clearpage
\appendix

\section{Simulation Platform Setup Details}
\label{app:setup}

\subsection{Duo-Wizard-of-Oz}
\label{app:duo-woz}

The majority of the VLN benchmarks separates language from demonstration in data collection, \textit{i.e.}, generate the trajectory first and then annotate them with language descriptions.
Our data collection, however, is inspired by the task-oriented in-vehicle dialogue corpora~\citep{kawaguchi2004ciair,hansen2005cu,eric2017key} collected through Wizard-of-Oz (WoZ) systems~\citep{riek2012wizard}.
In each gameplay session, a na\"ive participant, who is unaware of the \textbf{Wizard}, will communicate with the vehicle to visit goal locations specified in a storyboard.
Each task in the storyboard can potentially be changed due to unexpected alterations of the goal or environmental conditions on-the-fly.
In this way, \simulator~is unique in that it involves human subjects and data collection with unexpected events in naturalistic scenarios.
Compared to stage-wise data collection, WoZ studies ensure synchronous and natural human-agent interaction, leading to more realistic interaction~\citep{dahlback1993wizard}.

We further extend the single Wizard framework by introducing a pair of Wizards in the loop: a Collaborative Wizard (\textbf{Co-Wizard}) that serves the role of the original Wizard, and an Adversarial Wizard (\textbf{Ad-Wizard}) to control the environment and task interface to generate adversarial events on-the-fly. 
Without the Co-Wizard's and participant's awareness, the Ad-Wizard will challenge their collaboration by creating environmental changes and/or introducing task changes with appropriate context.
The Co-Wizard and participant need to communicate and negotiate to arrive at an alternative plan in order to address the unexpected situations and complete the navigation tasks.
The complete setup is supported by the Dialogue On the ROad To Handle Irregular Events (\simulator) platform we developed (illustrated in Figure~\ref{fig::simulator}).

\subsection{Framework and Notations}
\label{app:notation}

\begin{figure}[htp]
  \centering
  \includegraphics[width=0.5\textwidth]{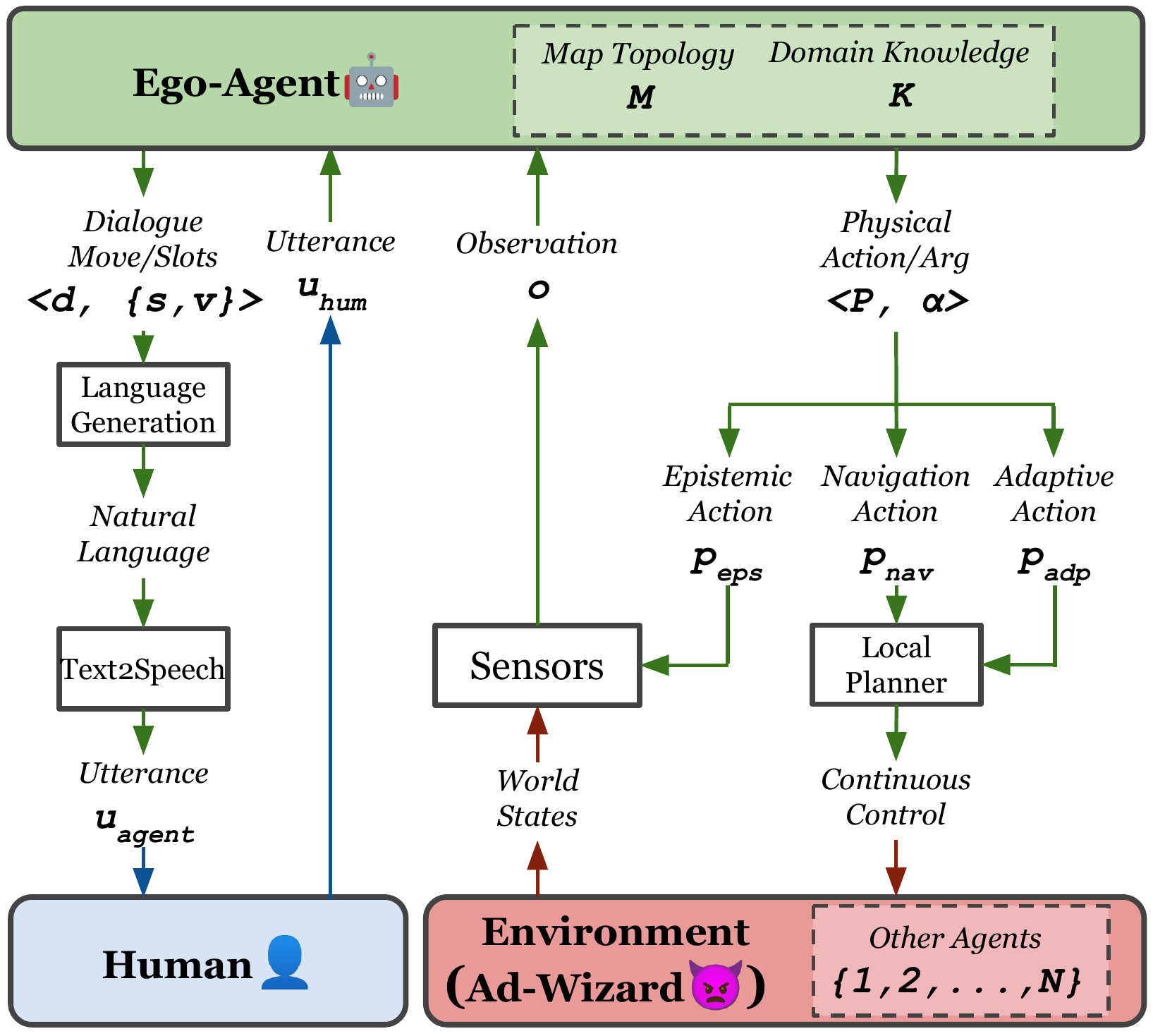}
  \caption{An overview of the framework of the \simulator~platform.}
  \label{fig::notation}
\end{figure}

The autonomous driving agent has a continuous control action space $\mathcal{A}$ as a triple over the normalized \textit{throttle} $\phi$, \textit{steering angle} $\theta$, and \textit{brake} $b$, which controls the vehicle in the simulated environment.
The agent is equipped with \textit{domain knowledge} $K$, including a list of \textit{street names} $\{\texttt{str}_i\}$ and (possibly incomplete) \textit{landmarks} $\{\texttt{lm}_i\}$ on the \textit{map} topology $M$.
At any time $t$, the \textit{interaction history} $H_t$ includes previous \textit{observation} and \textit{dialogue} $\{ O_{t-1}^H, \langle U_{t-1, \text{hum}}^H, U_{t-1, \text{agent}}^H \rangle \}$.
The agent is, in an end-to-end manner, a system that takes $H_t$ and \textit{action history} $A_t$ to produce an \textit{utterance} $u_{t,\text{agent}}$ and an \textit{action} $a_t \in \mathcal{A}$.
Due to the long-range driving and rich interactive context, training such an end-to-end policy is unrealistic~\citep{roh2020conditional}.
Following existing systems with a modular pipeline~\citep{skinner2016high,franke2017autonomous,Dosovitskiy17}, we break the problem down into \textit{perception}, \textit{planning}, \textit{control}, and \textit{interaction}, and summarize the framework in Figure~\ref{fig::notation}.

\paragraph{Navigation Actions}
To navigate, the agent has a discrete and finite physical action space $\mathcal{P} = \mathcal{P}_{\text{nav}} \cup \mathcal{P}_{\text{adp}} \cup \mathcal{P}_{\text{eps}}$.
Each \textit{navigation action} in $\mathcal{P}_{\text{nav}} = \{ \texttt{LaneFollow},$ $\texttt{LaneSwitch}, \texttt{JTurn}, \texttt{UTurn}, \texttt{Stop}, \texttt{Start} \}$ and its argument $\alpha \in \Theta$ is taken by the \textit{local planner} to produce the continuous action $a$.
The local planner defines a low-level policy $\pi \in \Pi: \mathcal{P}_{\text{nav}} \times \Theta \rightarrow \mathcal{A}$.

\paragraph{Adaptive Action}
An adaptive action $p \in \mathcal{P}_{\text{adp}}: \Pi \rightarrow \Pi$ adapts the low-level policy, \textit{e.g.}, by changing the target speed. 
In our case $\mathcal{P}_{\text{adp}} = \{ \texttt{SpeedChange} \}$, with an increment of 5.

\paragraph{Sensors and Epistemic Action}
The sensors of the agent defines an observation function $\omega \in \Omega: \mathcal{S} \rightarrow \mathcal{O}$ that maps the world state $s$ to an observation $o$ (in particular, an RGB image).
An epistemic action~\citep{kirsh1994distinguishing} is an action taken to facilitate mental computation instead of task completion, usually by manipulating sensors, \textit{e.g.}, selecting active sensor types and changing camera transforms.
It changes the observation function $\mathcal{P}_{\text{eps}}: \Omega \rightarrow \Omega$.
We assume for now a fixed first-person RGB camera with no noise, with only $\mathcal{P}_{\text{eps}} = \{ \texttt{LightChange} \}$.

\paragraph{Interaction}
To interact with human, the agent select a the dialogue move-slots pair $\langle d, \{s,v\} \rangle$, as described in Section~\ref{sec:annotate}.
The language generation module generates natural language as a sequence of tokens and produces the utterance $u_{\text{agent}}$ with Google Text-to-Speech (gTTS)\footnote{\url{https://gtts.readthedocs.io/}}.

\begin{figure*}[!htp]
    \centering
    \begin{subfigure}[t]{1.0\textwidth}
        \centering
        \includegraphics[width=0.95\linewidth]{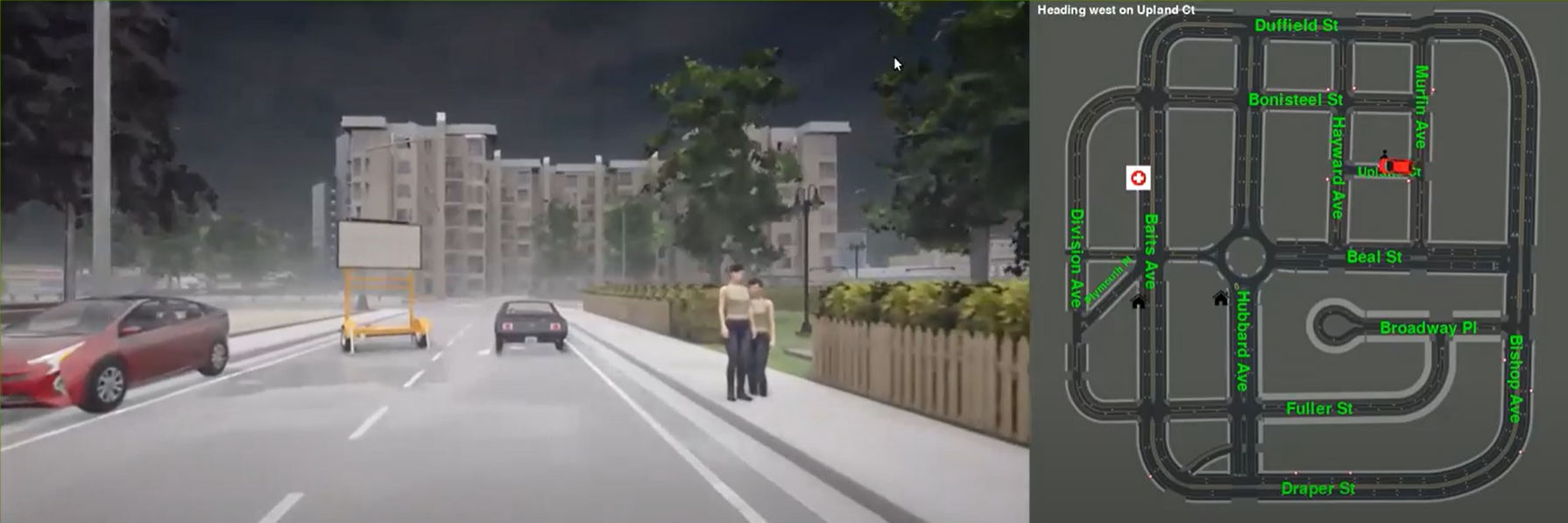}
        \caption{The participant's interface.}
        \label{fig::hum-interface}
    \end{subfigure}   
    ~ \\
    \begin{subfigure}[t]{1.0\textwidth}
        \centering
        \includegraphics[width=0.95\linewidth]{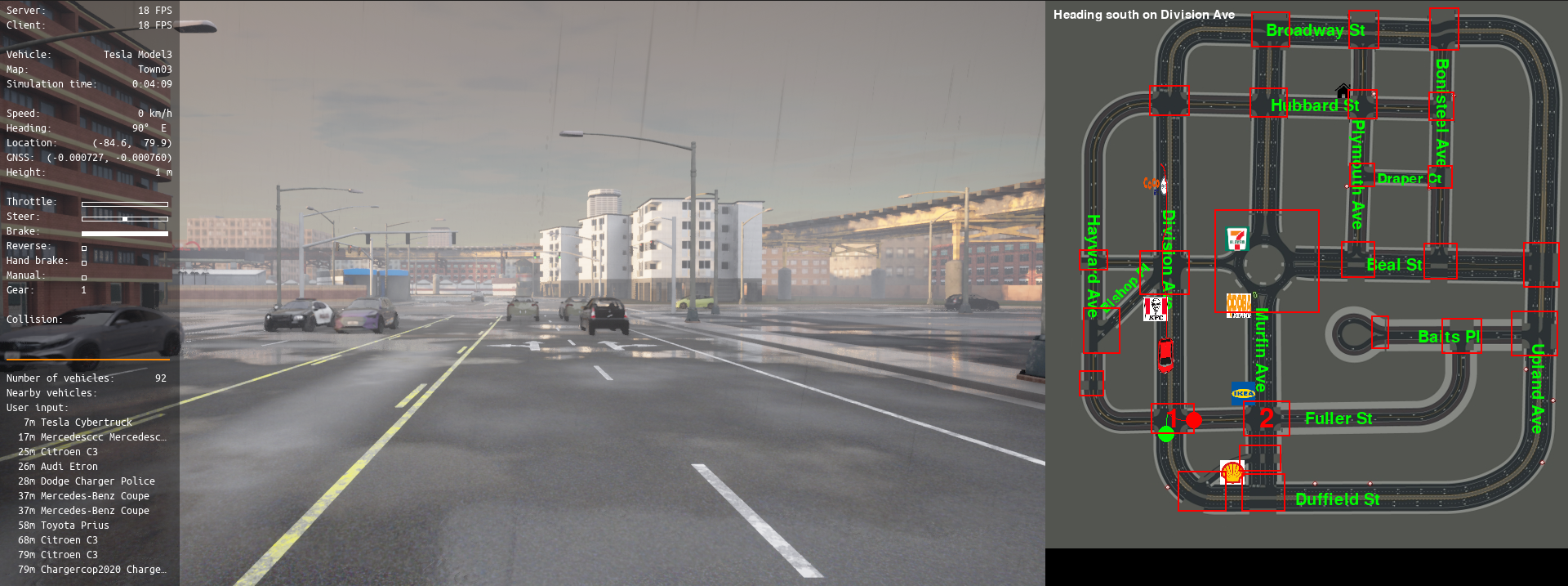}
        \caption{The Co-Wizard's interface.}
        \label{fig::wiz-interface}
    \end{subfigure}
    \vspace*{-0.25cm}
    \caption{Graphical interfaces for the participant and Co-Wizard.}
    \label{fig::interface}
    \vspace*{-0.5cm}
\end{figure*}

\subsection{Simulated Environment Details}
\label{app:carla}

\texttt{CARLA} supports waypoint precision of 2cm to benchmark the continuity of robotics navigation, as well as HD Maps under the OpenDrive 1.4 Standard~\citep{dupuis2006opendrive}, with unique IDs for lanes, roads, and junctions to represent high-level navigation plans.
To benchmark the rich environmental dynamics of outdoor navigation, we simulate multi-agent traffic with other active vehicles, bikers, and pedestrians, as well as different weather and light conditions in \texttt{CARLA}.

\paragraph{Multi-agent Environment}

Weather and light conditions can be controlled and configured on-the-fly. 
\texttt{CARLA} simulates multiple agents sharing the same environment, including vehicles and pedestrians. 
We use \texttt{CARLA}'s built-in traffic manager to simulate realistic traffic behaviour.

\paragraph{Multi-sensory Perception}

\texttt{CARLA} allows flexible sensor suites on agents, including realistic sensors (\textit{e.g.}, RGB, LIDAR, Radar, IMU, GNSS) and pseudo-sensors of ground truth (\textit{e.g.} depth, semantic segmentation, obstacle detector).
While our experiment involves only RGB images with object bounding boxes, depth, and semantic segmentation for supervision, it is possible to obtain additional sensory data for more complicated multi-modal studies.

\subsection{Navigation Task Setup}
\label{app:navigation}

\paragraph{Task Configuration}
In each trial, a storyboard will specify the names and contexts of two to six landmarks to visit.
Except for the final destination, the intermediate subgoals are unordered.
The participant and Co-Wizard need to collaborate to guide the vehicle through all the intermediate landmarks, starting from a departure location to a destination landmark.
Each session configuration is seeded from four different towns and a set of storyboard templates, with all landmark locations, street names and departure locations randomly shuffled.
An example storyboard template is presented as follows:

{\small
\begin{verbatim}
{
  "story": 
    "Your friend Annabel is moving to a new house, 
    and you decided to help her by doing some 
    shopping for her. You need to get $I1 and 
    $I2 from $P1 and $I3 from $P2, and head to 
    Annabel's new $P3 to help her clean the house.",
    
  "subgoals": [
    {"destination": "$P1", 
     "description": "Pick up $I1 and $I2 from $P1"},
    {"destination": "$P2", 
     "description": "Pick up $I3 from $P2"},
    {"destination": "$P3", 
     "description": "Arrive at Annabel's new $P3"}
  ],
  
  "variables": 
    [["P1", "places.stores"], 
     ["P2", "places.stores"], 
     ["P3", "places.residential"]],
     
  "dependents": 
    [["I1", "P1.items"], 
     ["I2", "P1.items"], 
     ["I3", "P2.items"]],
}
\end{verbatim}
}

\paragraph{Knowledge Disparity}
Both the Co-Wizard and the participant perceive the environment through a stream of RGB images. 
To replicate realistic outdoor navigation, an aerial map of the environment, with landmarks and current location, is provided to both the participant and the Co-Wizard.
While both players have access to some landmarks (\textit{e.g.}, the location of a restaurant or grocery store), the Co-Wizard does not have access to some of the landmarks (\textit{e.g.}, the location of a friend's house or a person to pick up).
Such knowledge disparities motivate situated communication beyond control asymmetry and challenge the agent to understand language instructions of different granularity.
For example, in the storyboard above, the Co-Wizard has no access to the location of Annabel's new house.

{\small
\begin{verbatim}
{
  "hidden_from_wizard": ["P3"]
}
\end{verbatim}
}

\subsection{Interface Components}
\label{app:interface}

Figure~\ref{fig::simulator} shows the conceptual overview of our interface design, including 4 major components: the Camera View, the Aerial View, the Task Interface, and the Communication Protocol. 
They are illustrated and described in Figure~\ref{fig::interface}.

\paragraph{Camera View}

Both the participant and the Co-Wizard have access to a first-person view of the simulated environment, similar to that of a driver.

\paragraph{Aerial View}

Both the participant and the Co-Wizard have access to a 2D aerial map of the town, with the location and heading direction of the ego-vehicle.
Various annotations are included to facilitate operation, including trajectory history, street names and landmarks.
The Co-Wizard is additionally shown the planned waypoints, possible trajectories at junctions, and vehicle status.

\paragraph{Task Interface}

The task interface is for human participants only.
It displays the landmarks to visit as specified by the storyboard template.
A subgoal is automatically fulfilled when the vehicle stops within 2 meters from the waypoint closest to the landmark, and the corresponding subgoal on the participant's interface will turn from white to green. 
The participant needs to communicate with the Co-Wizard in natural language to inform the status of a subgoal.
When a subgoal is added, changed, or deleted by the Ad-Wizard, the task interface will correspondingly change.

\paragraph{Communication Protocol}

Instead of typing on a keyboard, our Communication Protocol contains a dialogue interface and speech-text conversion to allow press-and-talk communication, requiring minimal distraction from operating the vehicle. 
This also allows the Co-Wizard to speak to the subject via speech synthesis to maintain the illusion of an automated operator to fulfill the WoZ purpose.

\begin{figure*}[!htp]
    \centering
    \hspace*{-1.0cm}
    \begin{subfigure}[t]{.3\textwidth}
        \centering
        \includegraphics[width=1.01\linewidth]{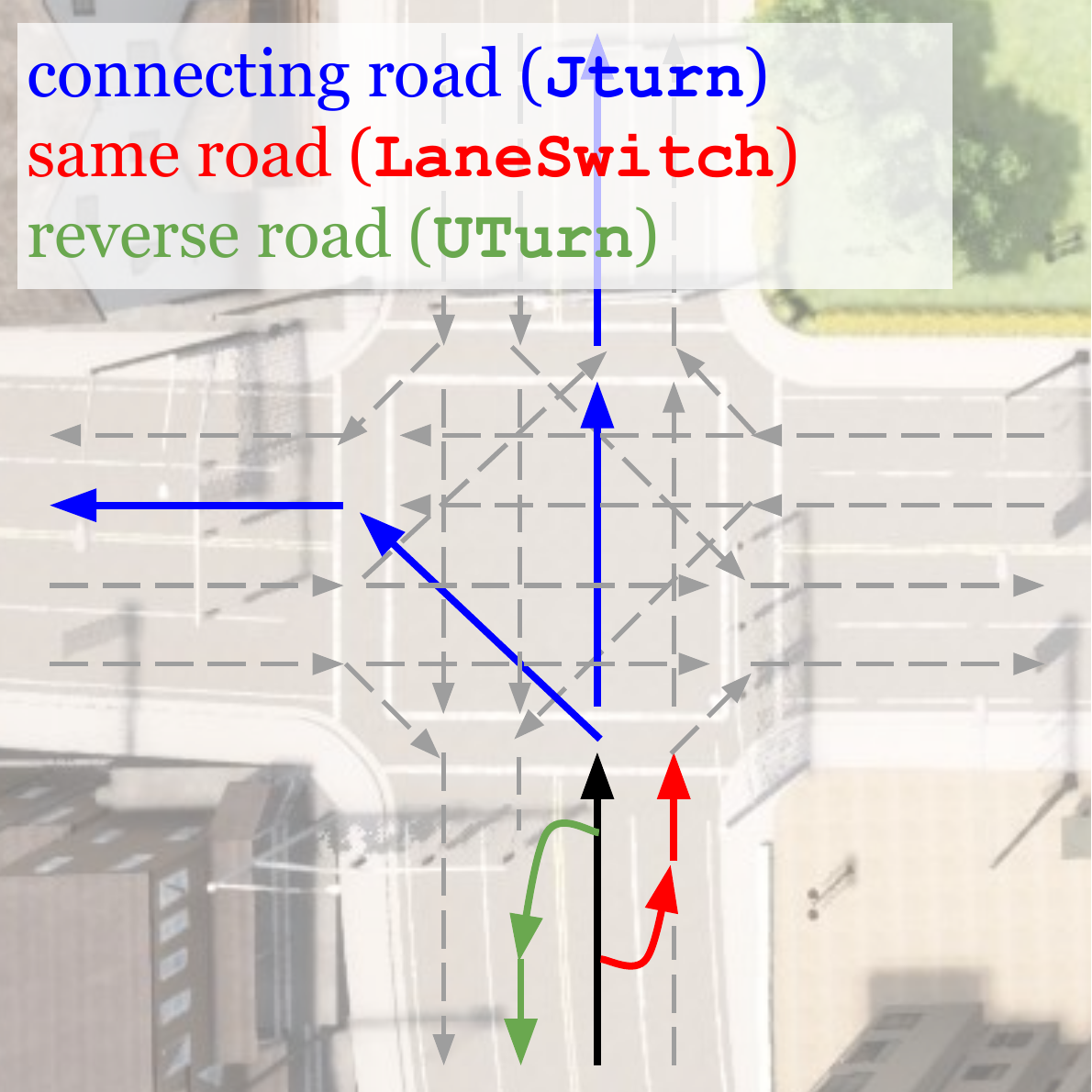}
        \vspace*{-0.25cm}
    \end{subfigure}   
    ~
    \begin{subfigure}[t]{.6\textwidth}
        \centering
        \vspace*{-4.9cm}
        \scalebox{0.6}{
    \begin{tabular}{lll}
            \toprule
            \textbf{Physical Actions} & \textbf{Args}      & \textbf{Descriptions}                 \\
            \midrule
            \texttt{LaneFollow}                & -                      & Default behaviour, follow the current lane. \\
            \texttt{LaneSwitch}                & Angle (\texttt{Rotation}) & Switch to a neighboring lane.               \\
            \texttt{JTurn}                     & Angle (\texttt{Rotation})                & Turn to a connecting road at a junction.    \\
            \texttt{UTurn}                     & -                      & Make a U-turn to the opposite direction.    \\
            \texttt{Stop}                      & -                      & Brake the vehicle manually.                \\
            \texttt{Start}                     & -                      & Start the vehicle manually.                \\
            \cdashlinelr{1-3}
            \texttt{SpeedChange}               & Speed ($\pm$5)        & Change the desired cruise speed by 5 km/h.  \\
            \texttt{LightChange}               & Light State (\texttt{On/Off})    & Change the front light state.     \\
            \midrule
            \textbf{Mental Actions} & \textbf{Args}       & \textbf{Descriptions}                                         \\
            \midrule
            \texttt{PlanUpdate}              & List[\texttt{Junction ID}]           & Indicate intended trajectory towards a destination.           \\
            \texttt{GoalUpdate}              & List[\texttt{Landmark}]              & Indicate current goal as an intended landmark.                \\
            \texttt{StatusUpdate}            & Tuple[\texttt{Landmark,Status}] & Indicate a change in task status. \\
            \texttt{KnowledgeUpdate}          & \texttt{x,y}                & Guess the location of an unknown landmark.      \\
            \texttt{Other}                   & -                       & Other belief state updates.                                   \\   
            \bottomrule 
    \end{tabular}
    }
    \end{subfigure}
    \vspace*{-0.5cm}
    \caption{The space of primitive physical actions and mental actions of \textit{Co-Wizard}.}
    \label{tab:action-full}
    \vspace*{-0.5cm}
\end{figure*}

\begin{figure*}[!htp]
    \centering
    \begin{subfigure}[t]{.32\textwidth}
        \centering
        \includegraphics[width=1.0\linewidth]{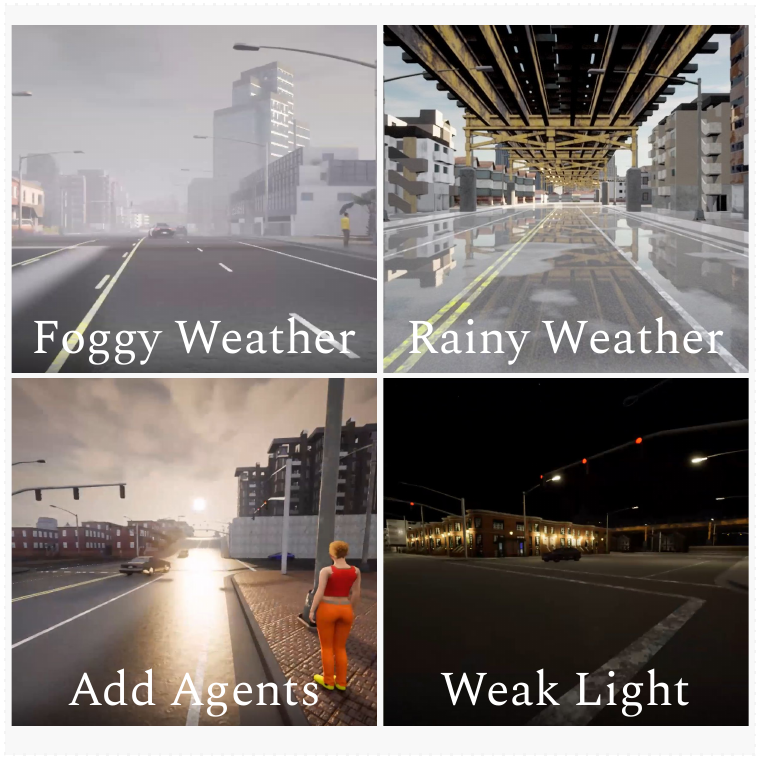}
        \vspace*{-0.5cm}
        \caption{Environmental Changes.}
        \label{fig::exception-env}
    \end{subfigure}   
    ~
    \begin{subfigure}[t]{.32\textwidth}
        \centering
        \includegraphics[width=1.0\linewidth]{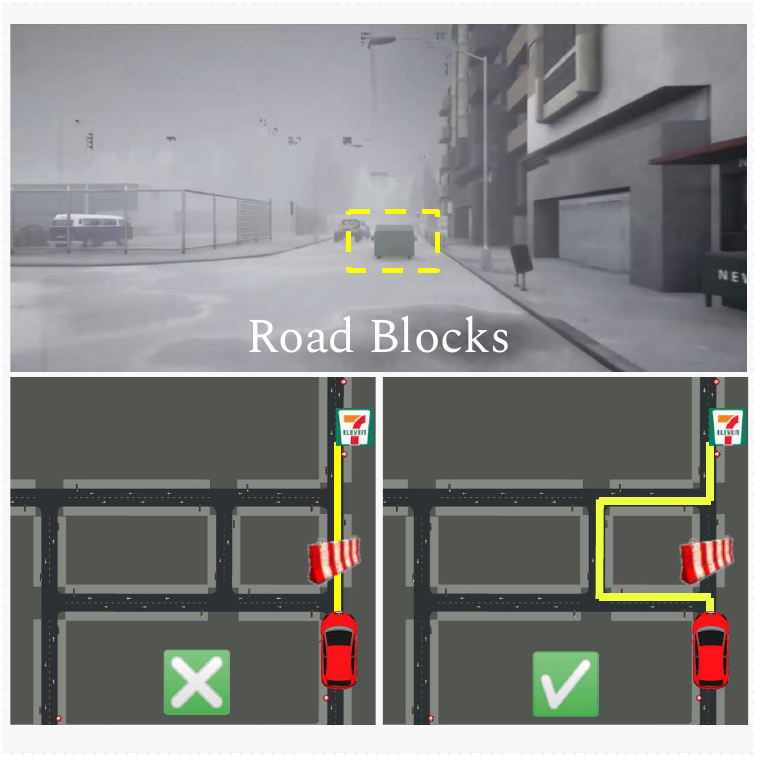}
        \vspace*{-0.5cm}
        \caption{Unexpected event at plan level.}
        \label{fig::exception-plan}
    \end{subfigure}
    ~    
    \begin{subfigure}[t]{0.32\textwidth}
        \centering
        \includegraphics[width=1.0\linewidth]{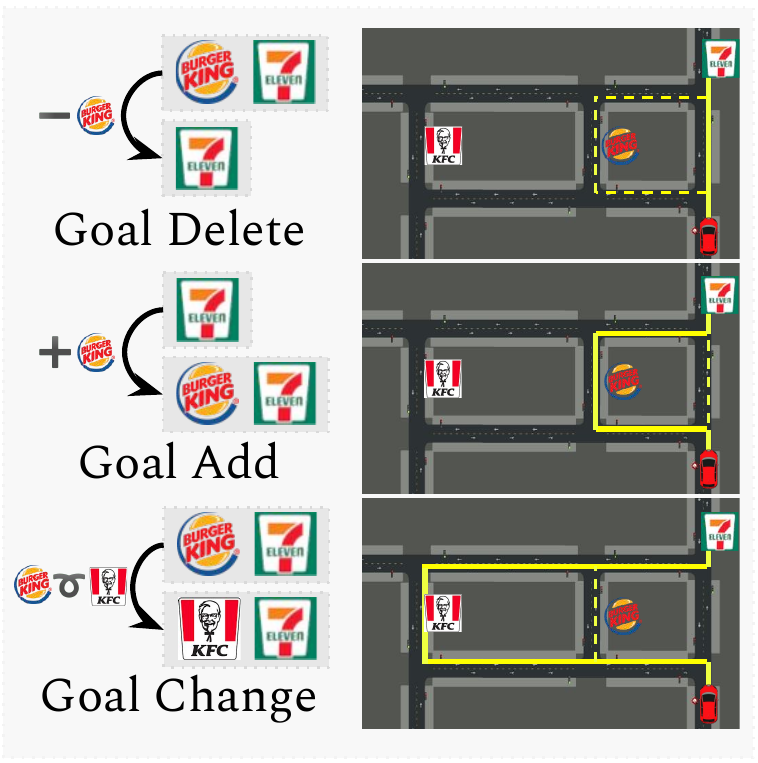}
        \vspace*{-0.5cm}
        \caption{Unexpected event at goal level.}
        \label{fig::exception-goal}
    \end{subfigure}
    \vspace*{-0.25cm}
    \caption{The Ad-Wizard is able to change the environment and tasks on-the-fly.}
    \label{fig::exception}
    \vspace*{-0.25cm}
\end{figure*}

\subsection{Co-Wizard Interface and Actions}
\label{app:cowiz-interface}

\paragraph{Physical Actions}
We found in pilot studies that the low-level free-form controller with continuous action space is not desirable, due to the poor quality of demonstrated trajectories and high cognitive load on the Co-Wizard.
On the other hand, a high-level, point-to-point maneuver controller~\citep{Dosovitskiy17} does not support the flexibility of human command, especially in our study with heavy replanning.
Motivated by prior work~\citep{roh2020conditional,codevilla2018end,muller2018driving}, we developed a set of high-level physical actions from pilot studies for the Co-Wizard to control the vehicle.
Similar to~\citet{muller2018driving}, we first map each action to a rule-based local trajectory planner to generate a list of waypoints that the vehicle will drive through, then feed the waypoints to a PID controller to produce the control signals for the agent.

\paragraph{Mental Actions}
In a complex navigation task with multiple subgoals, {\em belief tracking} over plans, goals, task status and knowledge becomes crucial~\citep{ma2012landmark,misu2014situated}.
Besides controlling the vehicle and communicating with the participant, the Co-Wizard also annotates the mental actions during and after the interaction.
\begin{itemize}
    \setlength\itemsep{-0.25em}
    \item \texttt{\textbf{PlanUpdate}}: Indicate a change in the intended trajectory towards a destination. This is done during interaction, by noting down the navigation plan by clicking junctions on the intended trajectory from current position to the destination.
    \item \texttt{\textbf{GoalUpdate}}: Indicate the current subgoal, \textit{i.e.}, an intended landmark. This is done during interaction, by clicking a known landmark on the aerial view interface.
    \item \texttt{\textbf{StatusUpdate}}: Indicate a change in the belief of the task status. Since the task interface is not available to the Co-Wizard, this is done post-interaction, by annotating which participant utterances indicate a change of task status.
    \item \texttt{\textbf{KnowledgeUpdate}}: Guess the location of an unknown landmark. Since some landmarks are hidden from the Co-Wizards, they need to guess where the destination is by comprehending the participant's descriptions. This is done during interaction, by clicking an arbitrary point on the aerial view interface. 
\end{itemize}

\subsection{Ad-Wizard Interface and Actions}
\label{app:adwiz-interface}

The Ad-Wizard is able to introduce {\em environmental exceptions} and {\em task exceptions}.
\begin{itemize}
    \setlength\itemsep{-0.25em}
    \item \textbf{Environmental Exceptions}: Triggered by the change to the environment. These include direct environmental changes (Figure~\ref{fig::exception-env}), which challenge the vehicle's perceptual processing and motivate participants to request for adaptations without changing the plan or goal (\textit{e.g.}, drive slowly in foggy weather and turn the headlights on at night). Environmental exceptions can also be introduced by creating road blocks (Figure~\ref{fig::exception-plan}), which motivate changes of plan by failing an original navigation plan towards a landmark.
    \item \textbf{Task Exceptions}: Brought by changing the tasks specified in the storyboard by deleting, adding, or changing a landmark to visit (Figure~\ref{fig::exception-goal}). The Ad-Wizard will send a message to prompt the participant in the message interface with appropriate context, and modify the task interface that specifies the landmarks to visit. Since the Co-Wizard does not have a task interface, the participant needs to communicate with the Co-Wizard in natural language to inform the status of a subgoal, especially when a change of current subgoal is indicated by the Ad-Wizard.
\end{itemize}

The Ad-Wizard is able to create environmental changes by modifying the weather and light conditions, and spawning more agents.
To motivate changes of plan on the fly, the Ad-Wizard can create roadblocks on the paths towards a destination.
Besides controlling the environment, the Ad-Wizard can also change the tasks specified in the storyboard by deleting, adding, or changing a landmark to visit.
For example, in Figure~\ref{fig::simulator}, the Ad-Wizard attempts to change the original plan by sending a text to the human subject.

\section{Dataset Details}
\label{app:data}

\subsection{Comparison of Settings}

We elaborate on our comparison of settings with existing language-conditioned navigation tasks in Table~\ref{tab:benchmarks} in the following dimensions:

\begin{itemize}
    \setlength\itemsep{-0.25em}
    \item \textbf{Environment Fidelity}: The environment in which the agent operates. Depending on the fidelity, the settings are categorized into \uline{Sim}ulated environment, \uline{Pano}ramic photos, and \uline{Phy}sical environment.

    \item \textbf{Environment Continuity}: Whether the environment is \uline{D}iscrete or \uline{C}ontinuous.

    \item \textbf{Turns of Communication}: Whether the communication between human and agent is \uline{S}ingle-term or \uline{M}ulti-term in a task.

    \item \textbf{Communication Form}: Whether the form of human-agent communication is in \uline{Freeform} Dialogue, \uline{Restricted} Dialogue that involves turn-taking QAs, or consists of \uline{Multi}ple \uline{Inst}ructions from the human only.

    \item \textbf{Language Granularity}: Whether the instructions are on the goal level (\uline{H}igh) or step-by-step on the movement level (\uline{L}ow).

    \item \textbf{Control Granularity}: Whether the actions are on the discrete action level (\uline{H}igh) or on the continuous control level (\uline{L}ow).

    \item \textbf{\uline{Lang}uage Collection}: Whether the language is collected from real \uline{H}umans or generated from pre-defined \uline{T}emplates.

    \item \textbf{\uline{Demo}nstration Collection}: Whether the demonstrations or trajectories are collected from real \uline{H}umans or generated from \uline{P}lanners.

    \item \textbf{Instruction Type}: The types of instructions initiated by a human, including possibly \uline{Replan}ning that requires a change of plan; \uline{Ad}a\uline{p}tation that requires adapting the manner of actions without changes of plans; \uline{Nav}igation that specifies a navigation action, or \uline{Mani}pulation that requests the agent to interact with an object.

    \item \textbf{\uline{Modal}ities}: The input modality to the agent's sensors, including possibly \uline{L}anguage (text), \uline{V}ision (images/videos), \uline{M}ap, or \uline{S}peech.
    
    \item \textbf{Action Space}: The output granularity of the agent's motors, possibly \uline{D}iscrete or \uline{C}ontinuous.
\end{itemize}

\subsection{Replay and Synchronization}
\label{app:replay}

We use \texttt{CARLA} 0.9.11 with \texttt{Unreal} 4.24 for data collection.
We apply asynchronously simulation at recording time for smooth interaction, and synchronous simulation at replaying time to retrieve sensory data at all frames without loss.
For each game, we record an interaction log and a game log under a fix time step of 30 FPS with 16 substeps for physics computation, and replay the session at 10 FPS following prior work~\citep{roh2020conditional}. 
The interaction log stores the timestamped activity history of the two wizards and the participant, including action history, spoken dialogue utterances, annotated mental actions, system prompts of the completion of tasks, etc.
The game log stores the world state at each timestamp to reproduce a game, including locations, orientations, bounding boxes, velocity, and physical control signals of vehicles and states of traffic lights, etc.
By attaching sensors to the ego-vehicle in replay, we are able to log RGB perception streams together with the ground truth pseudo-sensors (depth and semantic segmentation).

\subsection{Dialogue Annotation}
\label{app:dialogue}

\paragraph{Annotating Transcripts}
The trimmed audio clips are first sent to Google Speech Recognition\footnote{\url{https://pypi.org/project/SpeechRecognition/}} for raw text.
We then listen to each trimmed clip and type the ground truth transcripts.
Based on the ground truth transcripts, we further annotate each dialogue session using four levels of linguistic units, described as follows.

\paragraph{Annotating Transactions}

\textit{Transaction Units (TUs)} are sub-dialogues that starts when a task is initiated and ends when it is completed or abandoned. 
We observed in the corpus that the participant sometimes describes the next tasks to perform when current task is still ongoing, leading to small pieces of conversation standing alone from the major transaction unit for that subgoal.
Due to the addition, change, or deletion of a subgoal, some transactions are interrupted and continued afterwards (Figure~\ref{fig::task}).
Therefore, we assign each utterance to the subgoal it aims for, and one of the task status in \texttt{Ongoing}, \texttt{Complete}, \texttt{Abandoned}, \texttt{Pending} to each subgoal whenever there is a change specified by the utterance.

\paragraph{Annotating Exchanges}
\textit{Exchange Units (EUs)} are sequences of dialogue moves towards common ground. 
They starts with an initiating utterance that has a purpose (\textit{e.g.,} a question) and ends when the expectations are fulfilled or abandoned (\textit{e.g.,} an answer).
We observed in the corpus that some exchange units overlaps because the participant and the Co-Wizard spoke up at the same time.
This is particularly common when the Co-Wizard initiates a conversation asking for instructions and meanwhile the participant is giving the command.
The annotators are tasked to match each utterance to an exchange unit, represented by its initiating utterance.

\paragraph{Annotating Dialogue Moves}
\textit{Dialogue Moves} are sub-categories of dialogue acts that drive conversation and update domain-specific information state within an exchange. 
We follow the coding scheme of~\citet{carletta1997reliability} to represent dialogue moves as a decision tree, with a slight modification to adjust to our domain ontology, as presented in Figure~\ref{fig:dialogue}. 
The 14 dialogue moves, together with \texttt{Irrelevant}, specify the space of conversational action in the human-vehicle dialogue.
The annotators are tasked to first split each utterance into text spans that contain only one dialogue move, and then assign the move to the span following the decision tree.

\paragraph{Annotating Dialogue Slots}
\textit{Dialogue Slots} are parameters that further determine the semantics of dialogue moves. 
We consider 5 slot labels: \texttt{Action}, \texttt{Street}, \texttt{Landmark}, \texttt{Status}, \texttt{Object}.
For each slot label, the slot value belongs to a finite set of possible values defined by the domain ontology, \textit{e.g.}, the \texttt{Action} is specified by the physical action space, and the \texttt{Object} is specified by \texttt{CARLA}'s built-in visual semantics.
\begin{itemize}
    \setlength\itemsep{-0.5em}
    \item 10 values for \texttt{Action}: \texttt{Queried}, \texttt{Unknown}, \texttt{LaneFollow}, \texttt{LaneSwitch}, \texttt{JTurn}, \texttt{UTurn}, \texttt{Stop}, \texttt{Start}, \texttt{SpeedChange}, \texttt{LightChange}.
    \item 17 values for \texttt{Street}: \texttt{Queried}, \texttt{Unknown}, \texttt{Baits}, \texttt{Beal}, \texttt{Bishop}, \texttt{Bonisteel}, \texttt{Broadway}, \texttt{Division}, \texttt{Draper}, \texttt{Duffield}, \texttt{Fuller}, \texttt{Hayward}, \texttt{Hubbard}, \texttt{Murfin}, \texttt{Plymouth}, \texttt{Upland}, \texttt{Highway}.
    \item 12 values for \texttt{Landmark}: \texttt{Queried}, \texttt{Unknown}, \texttt{BurgerKing}, \texttt{Coco}, \texttt{Ikea}, \texttt{KFC}, \texttt{Panera}, \texttt{Qdoba}, \texttt{SevenEleven}, \texttt{Shell}, \texttt{House}, \texttt{Person}.
    \item 6 values for \texttt{Status}: \texttt{Queried}, \texttt{Unknown}, \texttt{Ongoing}, \texttt{Complete}, \texttt{Abandoned}, \texttt{Pending}.
    \item 24 values for \texttt{Object}: \texttt{Queried}, \texttt{Unlabeled}, \texttt{Building}, \texttt{Fence}, \texttt{Pedestrian}, \texttt{Pole}, \texttt{RoadLine}, \texttt{Road}, \texttt{SideWalk}, \texttt{Vegetation}, \texttt{Vehicles}, \texttt{Wall}, \texttt{TrafficSign}, \texttt{Sky}, \texttt{Ground}, \texttt{Bridge}, \texttt{RailTrack}, \texttt{GuardRail}, \texttt{TrafficLight}, \texttt{Static}, \texttt{Dynamic}, \texttt{Water}, \texttt{Terrain}, \texttt{Other}.
\end{itemize}

\paragraph{Kappa Analysis}

We report the average Cohen’s kappa ($\kappa$)~\citep{cohen1960coefficient} for each pair of annotators sharing common judgements in 20 common sessions with 1045 utterances.
We obtain $\kappa = 0.77\pm 0.02$ for dialogue move annotation, and $\kappa = 0.85\pm 0.02$ for dialogue slot annotation.

\subsection{Additional Dialogue Samples}
\label{app:sample}

The human and the agent communicate about objects in the scene, and the human frequently checks the perceptual capability of the agent.

\begin{figure}[H]
    \centering
    \includegraphics[width=0.5\textwidth]{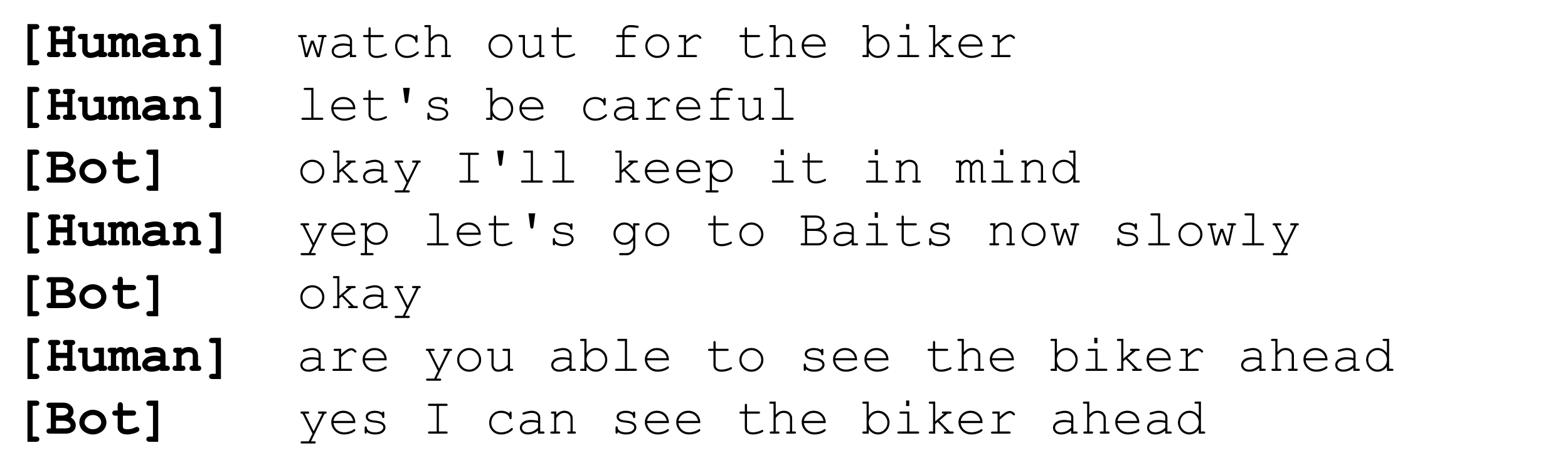}
\end{figure}

When a new task is introduced, the human and the agent negotiate towards a mutually agreed plan before the agent takes action.

\begin{figure}[H]
    \centering
    \includegraphics[width=0.5\textwidth]{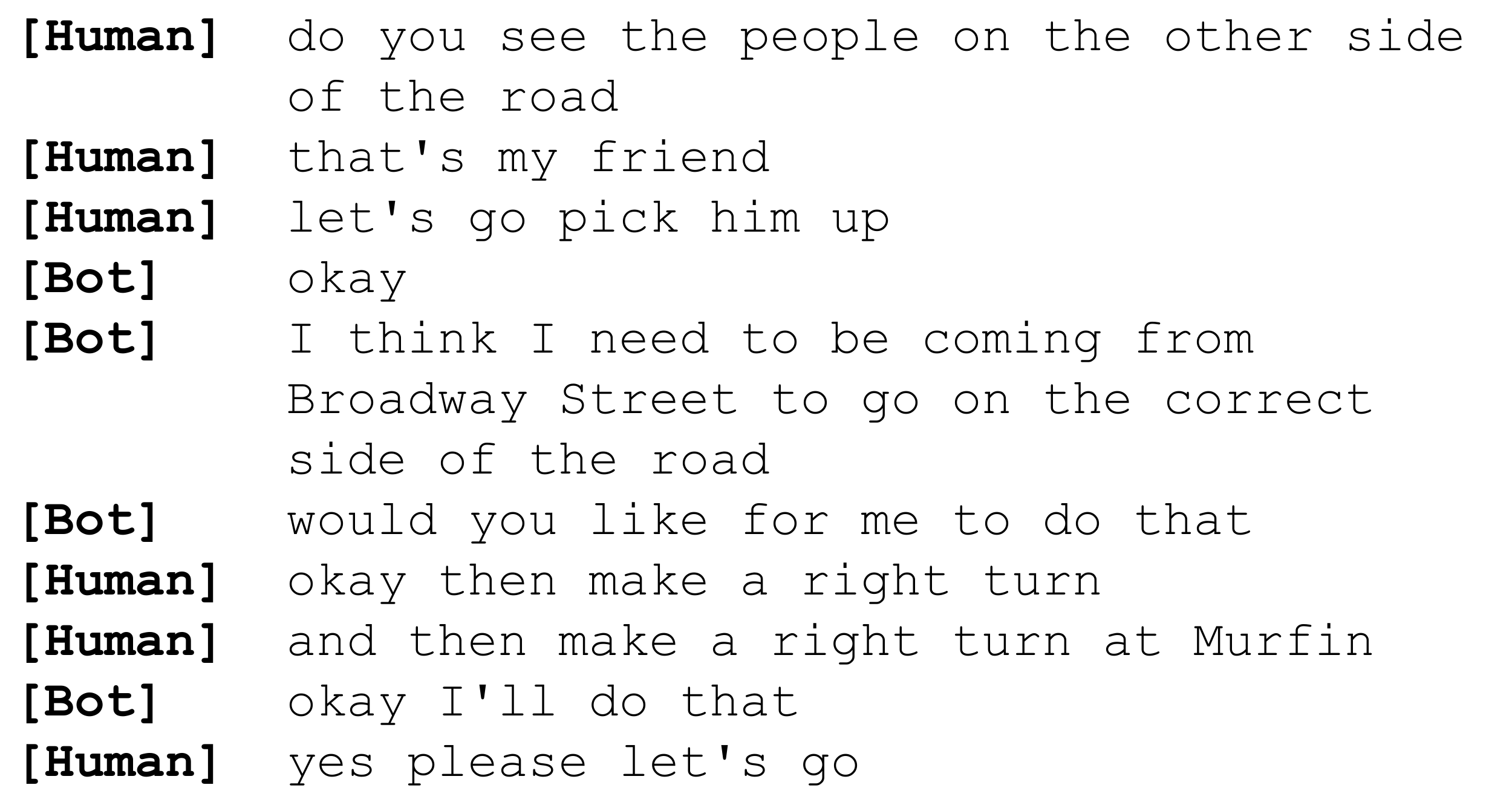}
\end{figure}

When the human gets confused by the agent's action, they would ask the agent to describe and explain its plan. This requires the agent to reason about the route planning task, and generate language description of it.

\begin{figure}[H]
    \centering
    \includegraphics[width=0.5\textwidth]{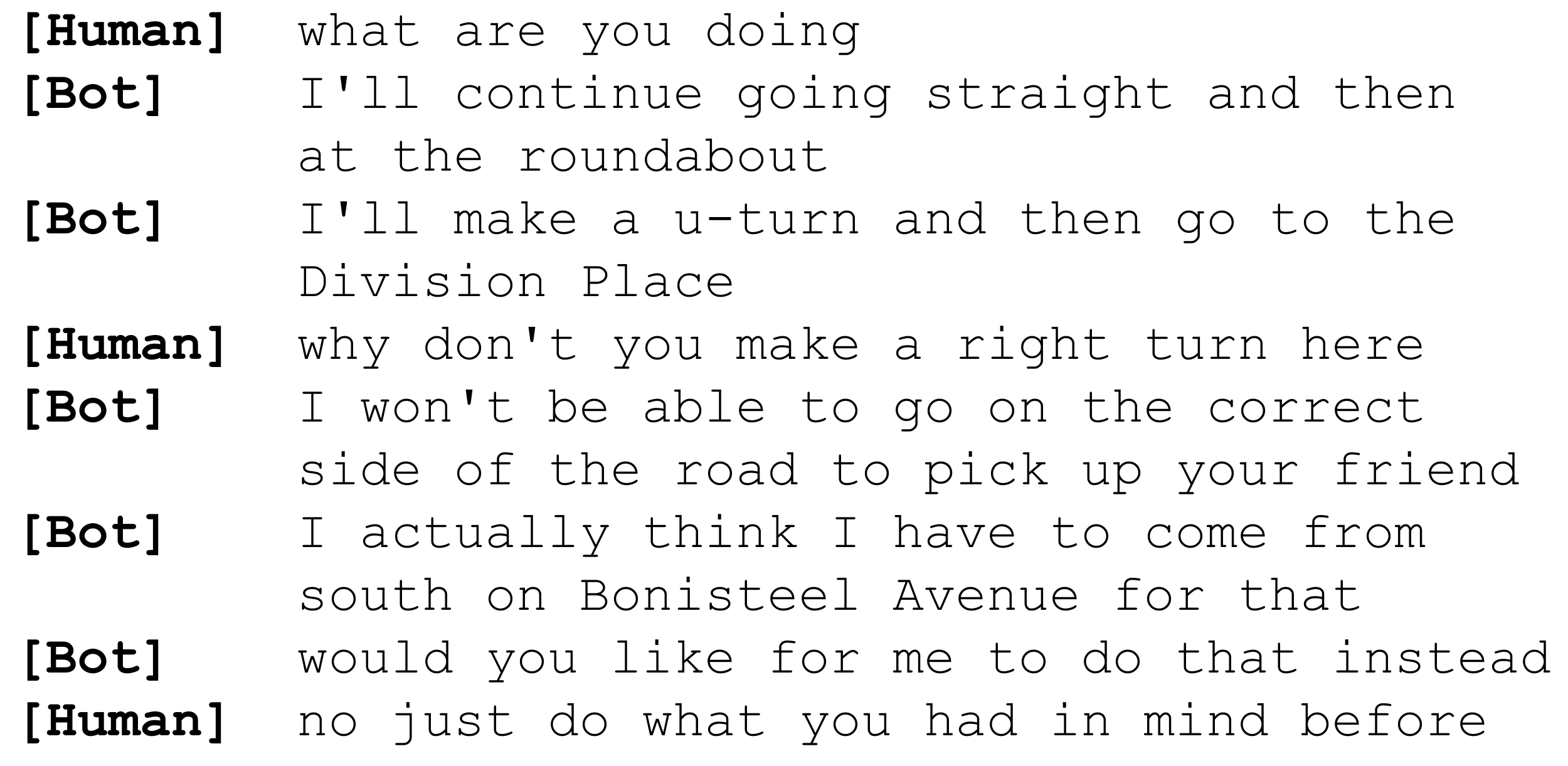}
\end{figure}

Episodic memory is also crucial to complete the tasks in~\dataset. 
The agent needs to keep track of the dialogue and visual history in order to resume a previously abandoned task.

\begin{figure}[H]
    \centering
    \includegraphics[width=0.5\textwidth]{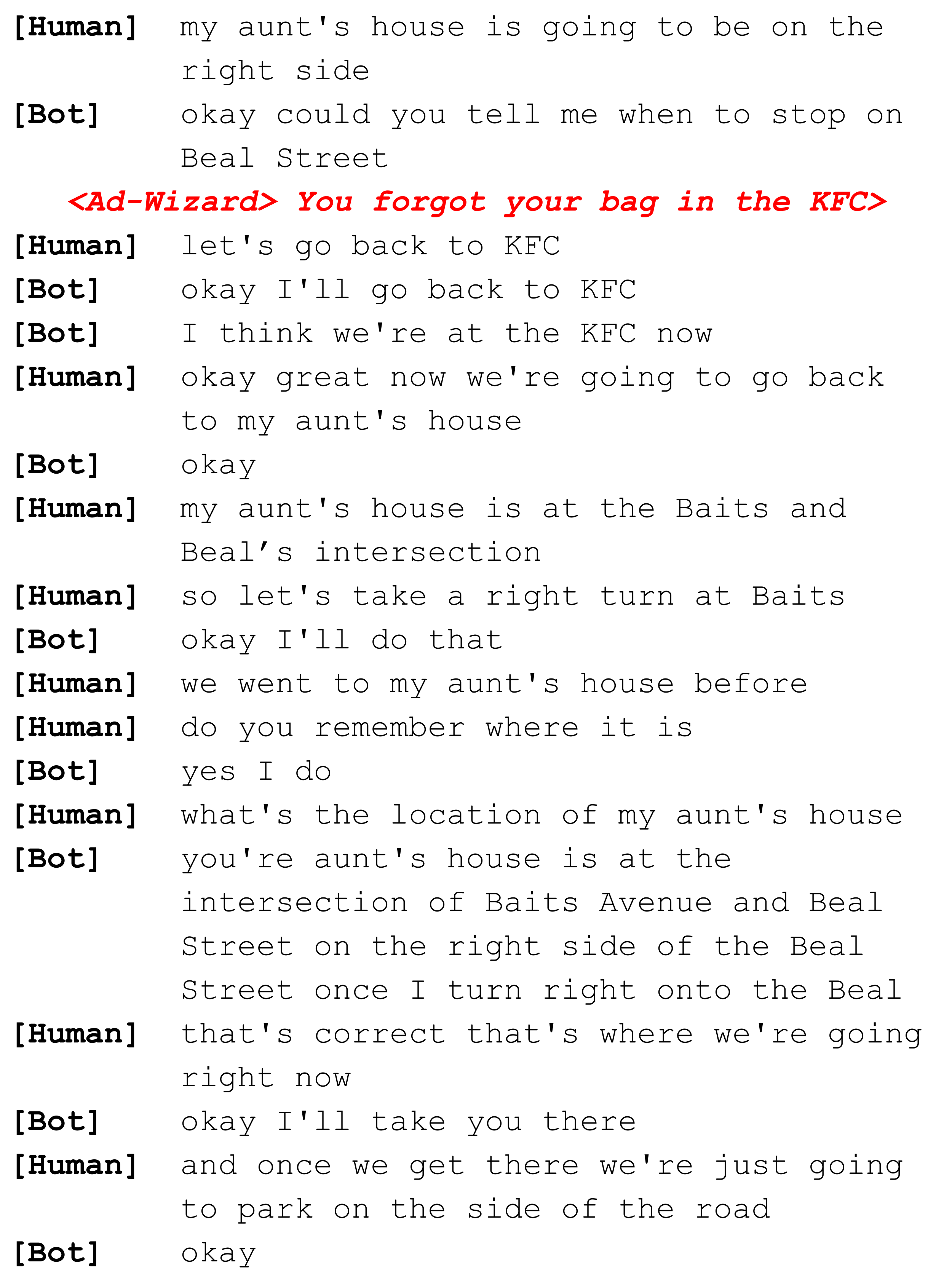}
\end{figure}
\section{Experiment Details}
\label{app:result}

\begin{figure*}[!htp]
    \centering
    \includegraphics[width=1.0\linewidth]{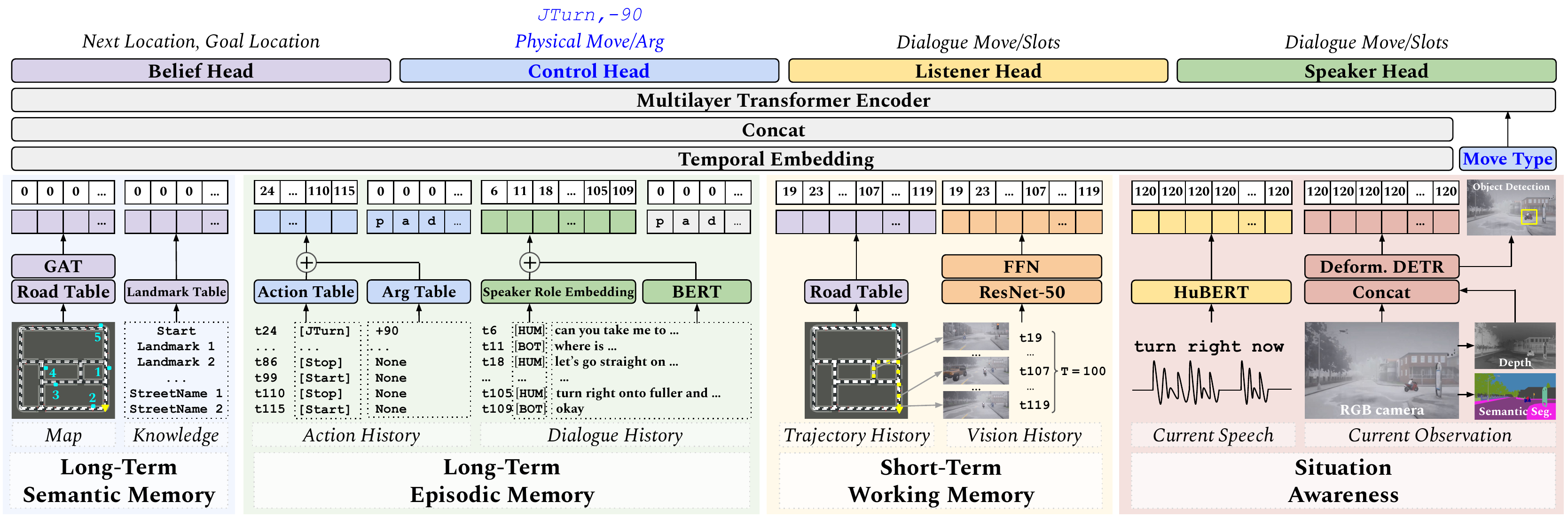}
    \caption{Temporally-Ordered Task-Oriented (TOTO) Transformer.}
    \label{fig::model-full}
\end{figure*}

\subsection{Computational Resources}

The recording of all sessions is done on a machine with an NVIDIA A4000 GPU.
The TOTO and baseline models are trained on one NVIDIA A40 GPU. 
For each experiment we train at least 10 epochs with early stop of 5 epochs, which take two hours in average.

\subsection{Baseline Details}
\label{app:baseline}

The complete model is illustrated in Figure~\ref{fig::model-full}.

\paragraph{Decision Transformer}
The encoding transformer has two layers each with 11 heads and dropout rate with 0.1. 
The length of input feature dimension is 770 and we use the sin-cos function along the whole game as temporal embedding of history. 
We used frozen BERT and HuBERT as our encoders for text and speech.
To process the image and extract features, we train a model using Segformer and Deformable-DETR. 
We first trained a Segformer model to predict depth and semantic segmentation using the RGB images generated by the simulator. 
Adding these two new channels into the RGB image, we augment the incoming image into 5 channels. 
We then develop a Deformable-DETR model with 300 object queries using the 5-channel images as input, and obtain a embedding with a size of $300\times256$ for each image.

\subsection{Hyper-Parameter Decisions}
We include major hyper-parameter decisions for reproducibility purposes. Please refer to supplementary code for more details.

\paragraph{Depth and Semantic Prediction (Segformer)}
With the same backbone as Segformer, another head is added to predict the depth along with the semantic segmentation.

\begin{itemize}
\setlength\itemsep{-0.25em}
    \item learning rate: $6\times10^{-5}$
    \item weight decay: $0.01$
    \item semantic class number: 23
    \item depth class number: 256
    \item optimizer: AdamW
\end{itemize}

\paragraph{Object Detection (Deformable-DETR)}
The original Deformable-DETR receives an RGB image input, while we have 5-channel (RGB, Depth, Semantics) augmented images.
We first use a fully connected layer to encode the input and then use the original Deformable-DETR model.
Two heads are used to predict the 2D and 3D bounding box for each query during training.
\begin{itemize}
\setlength\itemsep{-0.25em}
    \item learning rate: $2\times10^{-4}$
    \item backbone learning rate: $2\times10^{-5}$
    \item weight decay: $1\times10^{-4}$
    \item optimizer: AdamW
\end{itemize}

\paragraph{Learning Parameters}
\begin{itemize}
\setlength\itemsep{-0.25em}
    \item learning rate: $1\times10^{-5}$
    \item weight decay: $1\times10^{-4}$
    \item optimizer: AdamW
\end{itemize}

\paragraph{Loss coefficients}
\begin{itemize}
\setlength\itemsep{-0.25em}
    \item location belief MSE loss: $1\times10^{-3}$
    \item UfN dialogue move type cross entropy loss: $1$
    \item UfN dialogue slot type BCE loss: $1\times10^{-2}$
    \item UfN dialogue slot value BCE loss: $2\times10^{-3}$
    \item RfN dialogue move type cross entropy loss: $1$
    \item RfN dialogue slot type BCE loss: $1\times10^{-2}$
    \item RfN dialogue slot value BCE loss: $2\times10^{-3}$
    \item NfD Physical action type cross entropy loss: $1$
    \item NfD Physical action arg MSE loss: $1\times10^{-2}$
\end{itemize}

\subsection{Addendum to Results}
\label{app:results}

\paragraph{Vision Model Performance}

The performance of the vision models on semantic segmentation and object detection are provided in Table~\ref{tab:vision-model}.
While the performance itself is competent, the decision transformer fails to benefit from the vision representations.
This observation indicates that reasoning over cross-modal information may be a bottleneck for current end-to-end baselines.

\begin{table}[!htp]
\centering
\scalebox{0.75}{
\begin{tabular}{ccccc}
    \toprule
    \multicolumn{3}{c}{Semantic Segmentation} & \multicolumn{2}{c}{Object Detection} \\
    \cmidrule(lr){1-3} \cmidrule(lr){4-5}
    mIoU         & mAcc         & aAcc        & AP$_\text{(IoU=0.50)}$  & AP$_{\text{(IoU=0.50:0.95)}}$  \\
    65.60\%      & 71.60\%      & 97.10\%     & 71.40\%        & 41.30\% \\
    \bottomrule
\end{tabular}
}
\vspace*{-0.1cm}
\caption{The performance of the vision models on semantic segmentation and object detection.}
\label{tab:vision-model}
\vspace*{-0.1cm}
\end{table}

\paragraph{Additional Ablations}

We provide complete ablation results in Table~\ref{tab:results-full}.

\begin{table*}[!htp]
\centering
\scalebox{0.85}{
    \begin{tabular}{lcccccc}
    \toprule
    \multicolumn{1}{c}{\multirow{2}{*}{\textbf{Model}}} & \multicolumn{2}{c}{\textbf{UfN (Seen)}} & \multicolumn{2}{c}{\textbf{RfN (Seen)}} & \multicolumn{2}{c}{\textbf{NfD (Seen)}} \\
    \multicolumn{1}{c}{}                                & \textbf{Move Acc.}    & \textbf{Slot F1}       & \textbf{Move Acc.}    & \textbf{Slot F1}      & \textbf{Action Acc.}       & \textbf{Act-Arg Joint Acc.} \\
   \cmidrule(lr){1-1} \cmidrule(lr){2-3} \cmidrule(lr){4-5} \cmidrule(lr){6-7}
    TOTO                                               
    & $ 40.9 _{(\pm 3.9 )}$ & $ 36.9 _{(\pm  0.0 )}$  & $ 29.2 _{(\pm  0.7 )}$ & $ 55.7 _{(\pm  0.2 )}$ & $ 41.2 _{(\pm  2.5 )}$ & $ 36.0 _{(\pm  3.4 )}$ \\
    \cmidrule(lr){1-1} \cmidrule(lr){2-3} \cmidrule(lr){4-5} \cmidrule(lr){6-7}
    TOTO (+ Belief Tracking)                                   
    & $ 39.5 _{(\pm 2.2 )}$ & $ 37.0 _{(\pm  0.1 )}$  & $ 28.8 _{(\pm  0.9 )}$ & $ 55.7 _{(\pm  0.2 )}$ & $ 40.7 _{(\pm  3.6 )}$ & $ 34.0 _{(\pm  4.7 )}$\\
    TOTO (+ Fine-tuned BERT)
    & $ 38.4 _{(\pm 2.7 )}$ & $ 36.9 _{(\pm  0.0 )}$  & $ 27.8 _{(\pm  0.7 )}$ & $ 55.7 _{(\pm  0.0 )}$ & $ 43.6 _{(\pm  1.6 )}$ & $ 30.0 _{(\pm  4.1 )}$\\
    \cmidrule(lr){1-1} \cmidrule(lr){2-3} \cmidrule(lr){4-5} \cmidrule(lr){6-7}
    TOTO (- Action History)                             
    & $ 30.5 _{(\pm 1.5 )}$ & $ 36.9 _{(\pm  0.0 )}$  & $ 23.5 _{(\pm  1.7 )}$ & $ 55.7 _{(\pm  0.0 )}$ & $ 27.6 _{(\pm  2.8 )}$ & $ 24.6 _{(\pm  4.0 )}$  \\
    TOTO (- GT Transcript)                             
    & $ 39.8 _{(\pm 1.9 )}$ & $ 36.9 _{(\pm  0.1 )}$  & $ 29.2 _{(\pm  0.8 )}$ & $ 55.6 _{(\pm  0.1 )}$ & $ 40.4 _{(\pm  3.4 )}$ & $ 31.6 _{(\pm  4.3 )}$ \\
    TOTO (- Object Detection)                           
    & $ 42.5 _{(\pm 2.8 )}$ & $ 37.0 _{(\pm  0.2 )}$  & $ 30.4 _{(\pm  0.7 )}$ & $ 55.8 _{(\pm  0.1 )}$ & $ 39.2 _{(\pm  3.5 )}$ & $ 34.4 _{(\pm  5.8 )}$  \\
    TOTO (- Vision History)                             
& $ 41.9 _{(\pm 1.3 )}$ & $ 37.0 _{(\pm  0.2 )}$  & $ 29.1 _{(\pm  0.5 )}$ & $ 55.8 _{(\pm  0.2 )}$ & $ 42.0 _{(\pm  3.1 )}$ & $ 36.1 _{(\pm  4.0 )}$ \\
    TOTO (- Current Speech)                             
    & $ 35.1 _{(\pm 2.7 )}$ & $ 36.7 _{(\pm  0.5 )}$  & $ 29.9 _{(\pm  0.9 )}$ & $ 55.9 _{(\pm  0.2 )}$ & $ 39.7 _{(\pm  1.9 )}$ & $ 33.7 _{(\pm  3.0 )}$\\
    TOTO (- Map Knowledge)                             
    & $ 42.6 _{(\pm 1.2 )}$ & $ 36.9 _{(\pm  0.0 )}$  & $ 29.3 _{(\pm  0.9 )}$ & $ 55.8 _{(\pm  0.2 )}$ & $ 44.6 _{(\pm  3.3 )}$ & $ 39.1 _{(\pm  3.3 )}$ \\
    \cmidrule(lr){1-1} \cmidrule(lr){2-3} \cmidrule(lr){4-5} \cmidrule(lr){6-7}
    Fine-tuned BERT Only
    & $ 66.8 _{(\pm 2.0 )}$ & $ 24.9 _{(\pm  5.5 )}$  & $ 52.7 _{(\pm  1.0 )}$ & $ 46.0 _{(\pm  2.5 )}$ & $ 32.4 _{(\pm  1.2 )}$ & $ 16.2 _{(\pm  2.7 )}$\\
    BERT Only                   
    &  $ 52.1 _{(\pm 3.2 )}$ & $ 39.9 _{(\pm  1.3 )}$  & $ 52.3 _{(\pm  1.0 )}$ & $ 56.1 _{(\pm  0.3 )}$ & $ 30.4 _{(\pm  1.8 )}$ & $ 25.6 _{(\pm  2.8 )}$\\
    HuBERT Only                  
    & $ 35.7 _{(\pm 3.4 )}$ & $ 36.9 _{(\pm  0.0 )}$  & $ 24.5 _{(\pm  0.1 )}$ & $ 55.6 _{(\pm  0.1 )}$ & $ 31.2 _{(\pm  0.9 )}$ & $ 31.1 _{(\pm  0.9 )}$\\
    Deformable-DETR Only                    
    & $ 31.5 _{(\pm 0.0 )}$ & $ 36.9 _{(\pm  0.0 )}$  & $ 24.4 _{(\pm  0.0 )}$ & $ 55.7 _{(\pm  0.0 )}$ & $ 31.5 _{(\pm  0.6 )}$ & $ 31.5 _{(\pm  0.6 )}$\\
    Map Encoder Only  
    & $ 25.3 _{(\pm 2.5 )}$ & $ 36.9 _{(\pm  0.0 )}$  & $ 22.5 _{(\pm  1.6 )}$ & $ 55.7 _{(\pm  0.0 )}$ & $ 29.7 _{(\pm  0.6 )}$ & $ 28.4 _{(\pm  1.8 )}$\\
    \cmidrule(lr){1-1} \cmidrule(lr){2-3} \cmidrule(lr){4-5} \cmidrule(lr){6-7}
    E.T.
    & $ 36.6 _{(\pm 3.6 )}$ & $ 37.0 _{(\pm  0.2 )}$  & $ 29.4 _{(\pm  1.2 )}$ & $ 55.9 _{(\pm  0.2 )}$ & $ 40.0 _{(\pm  2.8 )}$ & $ 32.2 _{(\pm  4.0 )}$\\  
    E.T. (+ Fine-tuned BERT)
    & $ 33.6 _{(\pm 1.1 )}$ & $ 36.8 _{(\pm  0.1 )}$  & $ 26.5 _{(\pm  0.9 )}$ & $ 55.7 _{(\pm  0.0 )}$ & $ 38.0 _{(\pm  1.1 )}$ & $ 27.6 _{(\pm  6.2 )}$\\
    \midrule
    \multicolumn{1}{c}{\multirow{2}{*}{\textbf{Model}}} & \multicolumn{2}{c}{\textbf{UfN (Unseen)}} & \multicolumn{2}{c}{\textbf{RfN (Unseen)}} & \multicolumn{2}{c}{\textbf{NfD (Unseen)}} \\
    \multicolumn{1}{c}{}                                & \textbf{Move Acc.}    & \textbf{Slot F1}       & \textbf{Move Acc.}    & \textbf{Slot F1}      & \textbf{Action Acc.}       & \textbf{Act-Arg Joint Acc.} \\
    \cmidrule(lr){1-1} \cmidrule(lr){2-3} \cmidrule(lr){4-5} \cmidrule(lr){6-7}
    TOTO                                            
& $ 49.2 _{(\pm 3.0 )}$ & $ 26.2 _{(\pm  0.0 )}$  & $ 31.0 _{(\pm  1.7 )}$ & $ 54.0 _{(\pm  0.7 )}$ & $ 45.8 _{(\pm  3.8 )}$ & $ 41.1 _{(\pm  2.8 )}$\\
    \cmidrule(lr){1-1} \cmidrule(lr){2-3} \cmidrule(lr){4-5} \cmidrule(lr){6-7}
    TOTO (+ Belief Tracking)                                   
& $ 47.1 _{(\pm 3.5 )}$ & $ 26.2 _{(\pm  0.0 )}$  & $ 29.0 _{(\pm  2.0 )}$ & $ 53.7 _{(\pm  0.7 )}$ & $ 47.6 _{(\pm  4.5 )}$ & $ 38.8 _{(\pm  3.1 )}$\\
    TOTO (+ Fine-tuned BERT)
& $ 49.6 _{(\pm 0.9 )}$ & $ 26.2 _{(\pm  0.1 )}$  & $ 34.0 _{(\pm  2.1 )}$ & $ 54.8 _{(\pm  0.0 )}$ & $ 48.5 _{(\pm  4.6 )}$ & $ 36.2 _{(\pm  5.5 )}$\\
    \cmidrule(lr){1-1} \cmidrule(lr){2-3} \cmidrule(lr){4-5} \cmidrule(lr){6-7}
    TOTO (- Action History)                             
& $ 35.5 _{(\pm 3.2 )}$ & $ 26.1 _{(\pm  0.1 )}$  & $ 28.2 _{(\pm  3.9 )}$ & $ 54.8 _{(\pm  0.0 )}$ & $ 36.8 _{(\pm  0.8 )}$ & $ 36.0 _{(\pm  1.7 )}$\\
    TOTO (- GT Transcript)                             
& $ 46.7 _{(\pm 2.4 )}$ & $ 26.2 _{(\pm  0.0 )}$  & $ 31.6 _{(\pm  2.6 )}$ & $ 54.2 _{(\pm  0.8 )}$ & $ 46.2 _{(\pm  5.9 )}$ & $ 37.6 _{(\pm  6.9 )}$\\
    TOTO (- Object Detection)                           
& $ 50.0 _{(\pm 1.8 )}$ & $ 26.2 _{(\pm  0.1 )}$  & $ 32.7 _{(\pm  2.2 )}$ & $ 53.8 _{(\pm  1.2 )}$ & $ 45.7 _{(\pm  5.2 )}$ & $ 40.3 _{(\pm  5.4 )}$\\
    TOTO (- Vision History)                             
& $ 48.7 _{(\pm 2.3 )}$ & $ 26.2 _{(\pm  0.1 )}$  & $ 31.5 _{(\pm  2.9 )}$ & $ 54.3 _{(\pm  0.7 )}$ & $ 45.9 _{(\pm  4.2 )}$ & $ 42.3 _{(\pm  3.5 )}$\\
    TOTO (- Current Speech)                             
& $ 42.8 _{(\pm 2.5 )}$ & $ 25.8 _{(\pm  0.3 )}$  & $ 33.8 _{(\pm  1.4 )}$ & $ 55.1 _{(\pm  0.4 )}$ & $ 46.5 _{(\pm  4.9 )}$ & $ 39.4 _{(\pm  5.2 )}$\\
    TOTO (- Map Knowledge)                             
& $ 48.2 _{(\pm 1.0 )}$ & $ 26.2 _{(\pm  0.1 )}$  & $ 31.9 _{(\pm  1.2 )}$ & $ 54.9 _{(\pm  0.8 )}$ & $ 51.7 _{(\pm  3.4 )}$ & $ 46.0 _{(\pm  4.0 )}$\\
    \cmidrule(lr){1-1} \cmidrule(lr){2-3} \cmidrule(lr){4-5} \cmidrule(lr){6-7}
    Fine-tuned BERT Only 
& $ 67.2 _{(\pm 1.5 )}$ & $ 16.2 _{(\pm  3.5 )}$  & $ 57.0 _{(\pm  0.9 )}$ & $ 46.9 _{(\pm  2.2 )}$ & $ 37.1 _{(\pm  1.5 )}$ & $ 19.6 _{(\pm  3.6 )}$\\
    BERT Only                   
& $ 57.3 _{(\pm 2.1 )}$ & $ 31.7 _{(\pm  2.1 )}$  & $ 57.4 _{(\pm  1.3 )}$ & $ 55.9 _{(\pm  0.6 )}$ & $ 35.3 _{(\pm  3.6 )}$ & $ 30.1 _{(\pm  4.2 )}$\\
    HuBERT Only                  
& $ 40.9 _{(\pm 2.3 )}$ & $ 26.2 _{(\pm  0.0 )}$  & $ 30.2 _{(\pm  0.2 )}$ & $ 54.7 _{(\pm  0.1 )}$ & $ 36.7 _{(\pm  0.1 )}$ & $ 36.7 _{(\pm  0.2 )}$\\
    Deformable-DETR Only                    
& $ 37.6 _{(\pm 0.0 )}$ & $ 26.2 _{(\pm  0.0 )}$  & $ 30.7 _{(\pm  0.0 )}$ & $ 54.8 _{(\pm  0.0 )}$ & $ 36.9 _{(\pm  0.1 )}$ & $ 36.9 _{(\pm  0.1 )}$\\
    Map Encoder Only  
& $ 30.1 _{(\pm 9.9 )}$ & $ 26.2 _{(\pm  0.0 )}$  & $ 23.1 _{(\pm  10.6 )}$ & $ 54.8 _{(\pm  0.0 )}$ & $ 37.0 _{(\pm  0.1 )}$ & $ 37.0 _{(\pm  0.1 )}$\\  
    \cmidrule(lr){1-1} \cmidrule(lr){2-3} \cmidrule(lr){4-5} \cmidrule(lr){6-7}
    E.T.
& $ 45.1 _{(\pm 3.8 )}$ & $ 26.1 _{(\pm  0.1 )}$  & $ 33.4 _{(\pm  2.2 )}$ & $ 54.7 _{(\pm  0.8 )}$ & $ 46.6 _{(\pm  3.3 )}$ & $ 37.0 _{(\pm  5.9 )}$\\
    E.T. (+ Fine-tuned BERT)
& $ 41.8 _{(\pm 0.6 )}$ & $ 26.2 _{(\pm  0.0 )}$  & $ 34.1 _{(\pm  1.1 )}$ & $ 54.8 _{(\pm  0.0 )}$ & $ 44.5 _{(\pm  3.1 )}$ & $ 32.9 _{(\pm  7.3 )}$\\

    \bottomrule
    \end{tabular}
}
\vspace*{-0.1cm}
\caption{Complete experiment results of \texttt{TOTO} and the baselines.}
\label{tab:results-full}
\vspace*{-0.1cm}
\end{table*}

\section{Ethical Considerations}
\label{app:ethics}

\subsection{Consent Statement}
\label{app:consent}

You are invited to participate in a research study that intends to develop approaches to support language communication between humans and autonomous vehicles (AV). 
If you agree to be part of the research study, you will be asked to interact with a simulated AV in a virtual world to accomplish a set of tasks. Imagine you need to send your car out to do some errands. While the car is out, there may be some unexpected situations happening (e.g., a fallen tree blocks the road, you need to add a stop, etc.) You will need to communicate with your car in natural language to help the car deal with these exceptions and achieve the tasks. The study will last approximately an hour. The interaction between you and the car (i.e., speech/chat) and screen activities (i.e., the movement of the car in the virtual environment and its surroundings) will be recorded in a datafile. The data collected in this study will be analyzed and used for research purposes. No personally identifiable information besides the audio recording will be stored in the datafile.

\subsection{Debriefing Statement}
\label{app:debrief}

Earlier in the consent form, we informed you that you will be asked to interact with a simulated AV in a virtual world to accomplish a set of tasks. In actuality, the vehicle is not controlled by an algorithm, but by our research staff. The exceptions you encountered during the study were generated by our research staff on the fly. Unfortunately, due to the nature of this Wizard-of-Oz study, we could not provide you with all of these details prior to your participation. This ensures that your reactions in this study were spontaneous and not influenced by prior knowledge that you are interacting with another human player. We regret the deception in the study, but we hope you understand the reason for it. Knowing all of the study information, you may withdraw your data without penalty or loss of benefits to which you are otherwise entitled.

\subsection{Complete Ethical Statement}
\label{app:ethic-statement}

The \dataset~benchmark contains human generated data (speech and demonstrations).
The institution’s Institutional Review Board (IRB) considered this project exempt from ongoing review.
Preceding each study, a staff will describe the capability of the vehicle to the participant.
The participant would sign a consent form in which they were asked to communicate with a fully-functional autonomous vehicle in a simulated environment and navigate to specified landmarks.
After the last session, the staff representative would debrief the participant by introducing the Wizard-of-Oz nature of the study.
Participants would sign another debriefing form to indicate their voluntary agreement to participate in this study.
Since some acquaintance with the \simulator~interface is required to play the wizard role, we trained five research staff on both the Co-Wizard and Ad-Wizard interface and they alternated in the human studies.
The data collection among research staff and volunteers are in line with standard ethical practice.

\subsection{Broader Impact}

For broader social impact, \simulator~aims at empowering autonomous vehicles with the ability to harness human knowledge and expertise through dialogue, and enabling natural language communication and collaboration in tackling unexpected situations.
Since the dataset was developed from the simulator, the safety concerns are minimal. 


\end{document}